\documentclass{article}

\usepackage{microtype}
\usepackage{graphicx}
\usepackage{booktabs} % for professional tables

\usepackage{hyperref}
\usepackage{caption}

\usepackage[accepted]{icml2025a}

\usepackage{amsmath}
\usepackage{amssymb}
\usepackage{mathtools}
\usepackage{amsthm}

\usepackage[capitalize,noabbrev]{cleveref}

\theoremstyle{plain}

\theoremstyle{definition}

\theoremstyle{remark}

\usepackage[textsize=tiny]{todonotes}

\usepackage{url}
\usepackage{inconsolata}
\usepackage{tabularx}
\usepackage{listings}
\usepackage{tikz}
\usetikzlibrary{shapes.geometric, positioning, calc}
\usepackage{xcolor}
\usepackage[most]{tcolorbox}
\usepackage{subfiles}
\usepackage{subfig}
\usepackage{makecell}
\usepackage{pdflscape}
\usepackage{rotating}

\icmltitlerunning{Knowledge Distillation from LLMs for Household Energy Modeling}

\newtcolorbox[number within=chapter,]{mySystem}[3][]{
arc=2mm,
lower separated=false,
fonttitle=\bfseries,
colbacktitle=red!10,
coltitle=black!50!black,
enhanced,
attach boxed title to top left={xshift=0.5cm, yshift=-2mm},
colframe=red!50!black,
colback=red!4,
breakable, % Enable line breaking
listing only, % Use listings for proper formatting
listing options={
    basicstyle=\ttfamily\footnotesize, % Monospace font
    breaklines=true, % Allow breaking long lines
    breakatwhitespace=true, % Break at spaces
    numbers=none, % Disable line numbers
},
overlay={
\node[draw=red!50!black,thick,
fill= red!10,rounded corners=1mm, 
yshift=0pt, 
xshift=-0.5cm, 
left, 
text=red!50!black,
anchor=east,
font=\bfseries] 
at (frame.north east) {#3};},
overlay={
fill= red!10,rounded corners=1mm, 
yshift=+1.2mm,
xshift=-0.5cm, 
left, 
font=\bfseries] 
at (frame.north east) {#3};},
title=#2,breakable}

\newtcolorbox[]{myUser}[3][]{
arc=2mm,
lower separated=false,
fonttitle=\bfseries,
colbacktitle=blue!10,
coltitle=black!50!black,
enhanced,
attach boxed title to top left={xshift=0.5cm, yshift=-2mm},
colframe=blue!50!black,
colback=blue!4,
breakable, % Enable line breaking
listing only, % Use listings for proper formatting
listing options={
    basicstyle=\ttfamily\footnotesize, % Monospace font
    breaklines=true, % Allow breaking long lines
    breakatwhitespace=true, % Break at spaces
    numbers=none, % Disable line numbers
},
overlay={
\node[draw=blue!50!black,thick,
fill= blue!10,rounded corners=1mm, 
yshift=0pt, 
xshift=-0.5cm, 
left, 
text=blue!50!black,
anchor=east,
font=\bfseries] 
at (frame.north east) {#3};},
overlay={
fill= blue!10,rounded corners=1mm, 
yshift=+1.2mm,
xshift=-0.5cm, 
left, 
font=\bfseries] 
at (frame.north east) {#3};},
title=#2,#1,breakable}

\newtcolorbox[]{myAssistant}[3][]{
arc=2mm,
lower separated=false,
fonttitle=\bfseries,
colbacktitle=green!10,
coltitle=black!50!black,
enhanced,
attach boxed title to top left={xshift=0.5cm, yshift=-2mm},
colframe=green!50!black,
colback=green!4,
breakable, % Enable line breaking
listing only, % Use listings for proper formatting
listing options={
    basicstyle=\ttfamily\footnotesize, % Monospace font
    breaklines=true, % Allow breaking long lines
    breakatwhitespace=true, % Break at spaces
    numbers=none, % Disable line numbers
},
overlay={
\node[draw=green!50!black,thick,
fill= green!10,rounded corners=1mm, 
yshift=0pt, 
xshift=-0.5cm, 
left, 
text=green!50!black,
anchor=east,
font=\bfseries] 
at (frame.north east) {#3};},
overlay={
\node[draw=green!50!black,thick,
inner sep=2mm,
fill= green!10,rounded corners=1mm, 
yshift=+1.2mm,
xshift=-0.5cm, 
left, 
font=\bfseries] 
at (frame.north east) {#3};},
title=#2 \thetcbcounter,#1,breakable}

\begin{document}

\twocolumn[
\icmltitle{Knowledge Distillation from Large Language Models for Household Energy Modeling}

\icmlsetsymbol{equal}{*}

\begin{icmlauthorlist}
\icmlauthor{Mohannad Takrouri}{equal,yyy}
\icmlauthor{Nicolás M. Cuadrado}{yyy}
\icmlauthor{Martin Takáč}{yyy}
\end{icmlauthorlist}
  
\icmlaffiliation{yyy}{Machine Learning Department, Mohamed Bin Zayed University of Artificial Intelligence, Abu Dhabi, United Arab Emirates}

\icmlcorrespondingauthor{Mohannad Takrouri}{mohannad.takrouri@mbzuai.ac.ae}

% You may provide any keywords that you
% find helpful for describing your paper; these are used to populate
% the "keywords" metadata in the PDF but will not be shown in the document
\icmlkeywords{Knowledge Distillation, LLM, Demand Modeling, Prompt Engineering, }
\vskip 0.3in
]

% this must go after the closing bracket ] following \twocolumn[ ...

% This command actually creates the footnote in the first column
% listing the affiliations and the copyright notice.
% The command takes one argument, which is text to display at the start of the footnote.
% The \icmlEqualContribution command is standard text for equal contribution.
% Remove it (just {}) if you do not need this facility.

% \printAffiliationsAndNotice{}  % leave blank if no need to mention equal contribution
\printAffiliationsAndNotice{\icmlEqualContribution} % otherwise use the standard text.

\begin{abstract}
Machine learning (ML) is increasingly vital for smart-grid research, yet restricted access to realistic, diverse data—often due to privacy concerns—slows progress and fuels doubts within the energy sector about adopting ML-based strategies. We propose integrating Large Language Models (LLMs) in energy modeling to generate realistic, culturally sensitive, and behavior-specific data for household energy usage across diverse geographies. In this study, we employ and compare five different LLMs to systematically produce family structures, weather patterns, and daily consumption profiles for households in six distinct countries. A four-stage methodology synthesizes contextual daily data, including culturally nuanced activities, realistic weather ranges, HVAC operations, and distinct `energy signatures’ that capture unique consumption footprints. Additionally, we explore an alternative strategy where external weather datasets can be directly integrated, bypassing intermediate weather modeling stages while ensuring physically consistent data inputs. The resulting dataset provides insights into how cultural, climatic, and behavioral factors converge to shape carbon emissions, offering a cost-effective avenue for scenario-based energy optimization. This approach underscores how prompt engineering, combined with knowledge distillation, can advance sustainable energy research and climate mitigation efforts.\footnote{Source code is available at \url{https://github.com/Singularity-AI-Lab/LLM-Energy-Knowledge-Distillation}}
\end{abstract}

\section{Introduction}

High-quality and diverse household energy consumption and generation datasets are crucial for advancing ML-based smart grid technologies. They enable accurate load forecasting, reduce bias and overfitting, and help integrate distributed renewable systems more effectively. With access to various patterns from different demographics, climates, building types, and generation/storage configurations, ML models can learn nuanced demand behaviors, optimize charging and discharging strategies, and detect anomalies or faults more reliably. Maturing ML-based strategies in the energy grid enables the development of personalized energy services—ranging from tailored demand response initiatives to intelligent tariff designs—while fostering system resilience, fairness, and efficiency. Ultimately, datasets drive sustainable innovation and improve grid management efficiency.

{\bf Motivation:}
The increasing complexity of energy consumption modeling and forecasting, considering the uncertainty introduced by renewable sources, has required the use of advanced computational techniques that go beyond traditional statistical approaches \cite{krzywanskiAdvancedComputationalMethods2024}. Household energy modeling requires a comprehensive understanding of various factors, including household composition, weather conditions, energy assets, appliances, cultural norms, and socioeconomic dynamics. For example, a study analyzing human-building interactions in Qatar used machine learning techniques to explore how socioeconomic and behavioral dimensions influence energy consumption, highlighting the importance of cultural and demographic factors in the development of energy policies \cite{jabbarReshapingSmartEnergy2021}. Similarly, research on residential load patterns has shown significant correlations between between these factors, underscoring the need to incorporate these variables into accurate energy consumption models \cite{weiCharacterizingResidentialLoad2021}. Existing methods for household energy modeling often depend on historical data or simulations, which may lack the granularity or adaptability required to accurately represent diverse scenarios. For instance, black-box models, which are data-driven and utilize historical data related to building energy consumption, require high-quality datasets; missing data and errors can directly reduce their accuracy and generalization capabilities \cite{yuBuildingEnergyPrediction2022}. 

Large Language Models (LLMs) have emerged as powerful tools for synthesizing data and knowledge across diverse domains \cite{tanLargeLanguageModels2024}, and the energy domain is not an exception. After the introduction of the concept of LLM-based agents in the research community \cite{park2023generative}, a new way of exploiting the knowledge already memorized in large models emerged. Prompt engineering (designing structured, context-rich instructions that align with the desired outcomes \cite{adminDesigningEffectivePrompts2023}) plays a critical role in guiding LLMs to generate accurate and relevant outputs \cite{woodPromptEngineeringLarge2024}, especially when it comes to defining agents through contextual information. By leveraging multiple LLMs—including proprietary and open-weight architectures—this study addresses the challenges of scalability and contextualization in energy modeling. We systematically compare five different LLMs to assess their effectiveness in generating culturally and behaviorally grounded household energy datasets. Specifically, we aim to integrate cultural, seasonal, and behavioral contexts into household energy consumption patterns.

\textbf{Contributions:} 
In this work, we harness \emph{LLM-driven knowledge distillation} to generate culturally nuanced and seasonally adaptive household energy datasets for six distinct countries. By systematically encoding cultural norms, seasonal variations, and weekday/weekend distinctions into structured prompts, we produce \emph{realistic family schedules} that reflect the intricate interplay of behavioral and climatic factors. We evaluated five LLMs, identifying three that consistently yielded robust, context-aware responses. To further enhance realism and reduce computational overhead, we propose an alternative approach that integrates \emph{external weather datasets}, thereby bypassing intermediate modeling steps while ensuring physically consistent data inputs. Our framework not only enriches the fidelity of simulated consumption patterns but also offers a \emph{context-sensitive perspective} on household energy modeling—bridging the gap between static assumptions and dynamic, culturally informed simulations. These findings demonstrate LLM utility in advancing energy optimization, shape policy decisions, and support sustainable development initiatives globally.

\textbf{Paper Structure:} 
The remainder of this paper is organized as follows: Section~\ref{Sec-Literature} provides a literature review, situating this study within the broader context of household energy modeling and LLM applications. Section~\ref{Sec-Methodology} discusses the methodology, detailing the prompt engineering process and data generation workflows. Section~\ref{Sec-Results} presents the results and analysis, highlighting the cultural and contextual diversity captured in the generated data. Section~\ref{Sec-Potentials} explores potential applications of our approach, and Section~\ref{Sec-Conclusion} concludes with key findings and directions for future work.

\section{Literature Review}\label{Sec-Literature}

Household energy research has historically emphasized the role of demographic and cultural factors in shaping energy consumption patterns but has often failed to fully integrate these insights into methodological approaches, highlighting a gap between theory and practice \cite{crosbieHOUSEHOLDENERGYSTUDIES2006}. The study in \cite{moraEnergyConsumptionResidential2018} highlights the significant impact of occupant behavior, socio-economic characteristics, and preferences (e.g., set-point temperatures and ventilation durations) on residential energy consumption. It emphasizes that variations in energy demand are influenced more by contextual factors like family composition and dwelling usage patterns than by physical attributes such as house size or floor area.

Traditionally, energy researchers made efforts to enable access to relevant anonymized data through direct measurement of the variables of interest, in other words, conducting in-place studies for a determined time and gathering as much data as they can from real places using sensors. The rationale is that by enabling more data, several other lines of research would benefit from diversifying the cases they study. In \cite{cruz2024pattern}, the authors were interested in the pattern-driven behavior on the building side in a demand-side management scenario. For their particular contribution, they used a thorough dataset available through the U.S. Energy Information Administration (EIA) \cite{hronis2020housing}. However, before selecting that dataset, they conducted a review of all the available datasets that contain demand information with the desired granularity and detail. Although they narrowed down the list to 20, the datasets varied considerably--especially in countries like the USA, the UK, and India--and exhibited several sources of bias.

Understanding and optimizing household energy consumption requires multidisciplinary approaches that combine insights from energy modeling, behavioral science, and computational advances. In \cite{duApplianceCommitmentHousehold2011}, the authors present an innovative algorithm that schedules household appliances based on consumption forecasts and user comfort settings, aiming to optimize objectives like minimizing costs or maximizing comfort. This framework underscores the importance of accurately modeling intra-family dynamics to effectively simulate appliance usage patterns. Building upon this, \cite{NICHOLLS2015116} examines how family interactions, such as shared activities and the presence of children, influence peak electricity demand times. The study highlights the necessity of incorporating family dynamics into energy consumption models to accurately predict and manage peak demand periods. Several studies have leveraged weather simulation frameworks to assess HVAC performance under various climatic scenarios. For example, \cite{xuEffectsCoolMaterials2024} analyzed the impact of urban heat islands on residential cooling demand, showing how microclimatic factors intensify energy consumption in densely populated regions. 

{\bf LLMs and Prompt Engineering:} LLMs such as OpenAI’s GPT and Meta’s Llama have transformed how domain-specific tasks are approached, particularly in generating human-like, context-aware outputs \cite{busterSupportingEnergyPolicy2024}. In the energy domain, these models have been employed to automate processes that previously required expert curation. For instance, \cite{busterSupportingEnergyPolicy2024} introduced a novel framework combining LLMs with decision-tree logic to analyze zoning laws for renewable energy siting, achieving high accuracy while enabling large-scale policy research. Additionally, \cite{majumderExploringCapabilitiesLimitations2024} explored the potential of LLMs in energy sector applications, highlighting opportunities for their integration into power system operations and optimization tasks. In \cite{almashor2024can}, the authors explored whether LLM-based agents can synthesize household energy consumption building upon the environment provided by \cite{park2023generative}. Although they obtained some reasonable load patterns, they were limited by the capabilities of the simulation where they ran their experiment; they limited their work to parsing keywords in the agent conversations and associating them to energy consumption to later sum the household load.

Recent studies highlight the potential of large language models to streamline data access and policy analysis in household energy modeling. However, significant gaps persist in integrating these components into a cohesive framework. Current models often rely on limited assumptions about family structures and behaviors, neglecting the nuanced and culturally specific dynamics necessary for accurate simulations. While weather data is commonly used in HVAC modeling, it is rarely contextualized to align with diverse family activities, leaving room for more integrated and adaptive approaches. To address these limitations, this research leverages LLMs to generate a consistent framework that considers diverse family structures and associated behaviors across countries, introducing a structured, multi-stage process for weather data generation. By identifying realistic ranges for key parameters such as temperature, humidity, solar radiation, and wind speed for each country and season, the methodology dynamically generates 24-hour weather profiles that reflect both macro-level seasonal trends and micro-level hourly variations.

\section{Methodology}\label{Sec-Methodology}

This work builds on three key themes: household energy modeling, weather data integration, and applications of large language models (LLMs) in domain-specific research. Prompt engineering plays a pivotal role in distilling knowledge from the LLMs. We guided our approach with the best practices proposed in \cite{woodPromptEngineeringLarge2024}: \textbf{(1) Contextual Specificity}: Prompts were designed to provide detailed context, ensuring the LLM understood the cultural, seasonal, and household-specific nuances, \textbf{(2) Iterative Refinement}: Prompts were iteratively refined based on initial outputs to improve clarity, accuracy, and adherence to desired formats. \textbf{(3) Structured Formats}: Outputs were requested in JSON or structured text formats, enabling seamless integration with downstream processing workflows, and \textbf{(4) Validation and Debugging}: Outputs were rigorously validated to ensure consistency and realism, with corrective adjustments made to prompts when necessary. 

\begin{table*}[t]
\centering
\caption{Summary of the five LLMs tested, including parameter sizes, and brief notes on performance. All models were accessed via DeepInfra.}
\label{tab:llms}
\begin{tabular}{ccc}
\toprule
\textbf{MODEL NAME (ID)} & \textbf{PARAMETERS} & \textbf{PERFORMANCE NOTES} \\
\midrule
\texttt{Llama-3.1-405B-Instruct} & 405B & \makecell{Best performance with stable, context-rich replies.} \\
\texttt{Llama-3.3-70B-Instruct} & 70B & \makecell{Accurate in nuanced prompts; good balance of speed and depth.} \\
\texttt{microsoft/phi-4} & 14B & \makecell{Produced context-aware outputs.} \\
\texttt{DeepSeek-R1} & 671B & \makecell{Occasional inconsistencies, incomplete reasoning and timeouts.} \\
\texttt{Qwen/QwQ-32B-Preview} & 32B & \makecell{Incomplete/circular reasoning; sometimes refused \\prompts previously answered.} \\
\bottomrule
\end{tabular}
\end{table*}
\begin{figure}[t]
    \centering
    \includegraphics[width=0.90\linewidth]{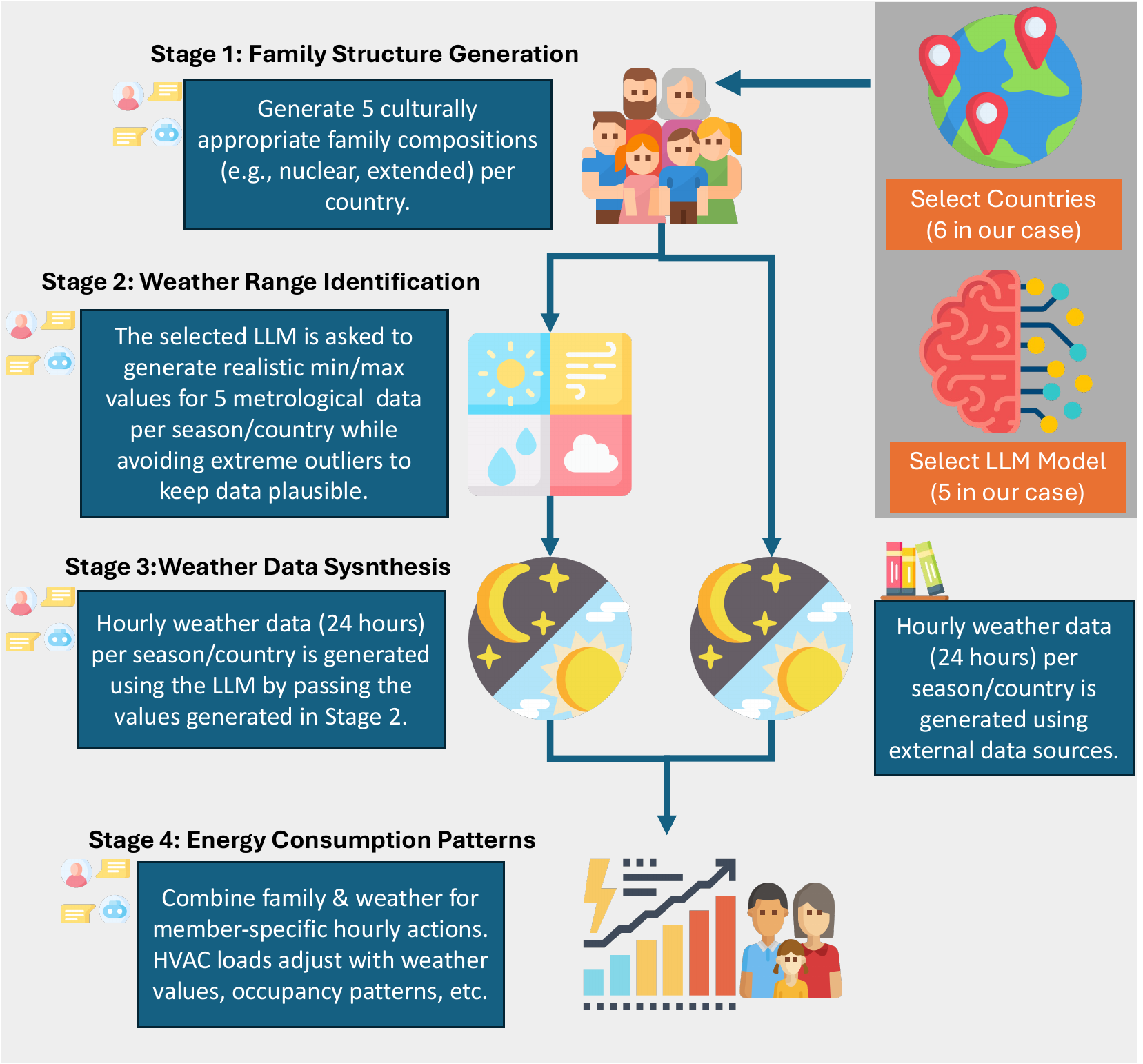}
    \caption{Overview of our framework with detail on the prompting strategy for each stage.}
    \label{fig:visual_abstract}
\end{figure}

We adopt a \emph{four-stage process} to leverage LLMs for generating synthetic data in household energy modeling, with each stage focusing on a critical aspect of the overall framework: \textbf{Stage 1:} Generate culturally specific family structures for six countries. \textbf{Stage 2:} Identify realistic ranges for key weather parameters  tailored to each country and season. \textbf{Stage 3:} Synthesize hourly weather data using the identified ranges, aligned with seasonal and geographical contexts. \textbf{Stage 4:} Model daily energy consumption patterns, incorporating member-specific activities and HVAC operations for various household types. These daily patterns are then aggregated into yearly profiles, incorporating country-specific weekends and holidays.\autoref{fig:visual_abstract} provides an overview of our methodology, highlighting the prompting sequence within each defined stage. 

Our primary goal is to leverage LLM-driven knowledge distillation to synthesize culturally and contextually rich data for household energy modeling. The internal workflow proceeds as follows: \textbf{Input Preparation:} Define the target countries, seasons, number of household types per country, and weather parameters of interest. Select the LLM to be used and decide whether weather data will be generated by the LLM (in the second and third stages) or sourced externally. \textbf{Prompt Execution:} Use the specified system and user prompts to guide the LLM in producing context-aware outputs across the different stages. \textbf{Data Synthesis:} Generate the results in structured formats such as JSON and CSV. \textbf{Visualization:} Develop interactive plots and dashboards to facilitate data exploration and analysis.

\subsection{LLMs Overview and Selection}
We tested five LLMs (\autoref{tab:llms}), accessed via DeepInfra \cite{MachineLearningModels}. Although all models produced culturally relevant text, \texttt{DeepSeek-R1} and \texttt{QwQ-32B-Preview} had inconsistencies/timeouts, so we used the first three models primarily for Stage~4. The following subsections detail our data-generation processes and workflows. Refer to Appendix~\ref{prompts} for all used prompts.

\subsection{Stage 1: Family Structures Generation}
The first stage focuses on creating culturally relevant family structures for six countries: the USA, Japan, India, Sweden, the United Arab Emirates, and Brazil. These countries were chosen to represent diverse cultural, geographic, and socioeconomic contexts. Prompts were crafted to elicit family compositions aligned with regional norms and practices. 

\textbf{Prompt Design:} 
A system prompt emphasized cultural sensitivity, diversity, and realism. It specified various family types (e.g., nuclear, extended, joint) to match typical household structures in each country. The user prompt requested five unique family types per country, formatted in structured JSON.

\subsection{Stage 2: Weather Parameter Ranges}
Stage~2 synthesizes realistic weather parameter ranges (temperature, humidity, solar radiation, wind speed) per season per country. Prompts were designed to generate seasonally appropriate minimum and maximum values, ensuring alignment with regional climatic norms. The output was formatted as structured text, capturing the dynamic weather ranges necessary for subsequent stages.

\textbf{Prompt Design:} The prompts explicitly requested seasonal variations for weather parameters, specifying realistic minimum and maximum ranges while avoiding extreme outliers. Temperature ranges are defined based on the seasonal climate of each country, ensuring alignment with observed averages and extremes. Solar radiation ranges are capped at 1000 W/m², with diffuse solar radiation restricted to remain below direct solar radiation during sunny hours.

\subsection{Stage 3: Detailed Weather Data Generation}
In the third stage, hourly weather data were generated for each country/season based on the values from Stage 2.

\textbf{Prompt Design:} 
The system prompt guided the LLM to vary meteorological conditions realistically throughout each 24-hour period, incorporating seasonal and geographic influences. The user prompt specified the country and season, requesting 24-hour weather data in structured form. For each country, the assistant’s response for one season served as input to the next season’s prompt, ensuring a consistent transition across seasons. We store results in CSV, capturing both macro-level seasonal trends and micro-level hourly changes for integration in subsequent modeling.

\subsubsection*{Alternative Approach: External Weather Data}
While the second and third stages rely on LLM-generated weather ranges and hourly data, we also explored an alternative approach that bypasses those stages by integrating \emph{externally sourced} Typical Meteorological Year (TMY) datasets. We fetch TMY data for each capital city using \texttt{pvlib} \cite{anderson2023pvlib}, yielding physically consistent hourly profiles in a matching CSV format, allowing for a seamless transition to Stage~4 regardless of the data source.

\subsection{Stage 4: Consumption Patterns and Yearly Profiles}
Stage~4 assigns 24 hourly actions and energy values to each household member, factoring in HVAC linked to weather data. Prompts emphasize cultural, seasonal, and weekday/weekend differences. For instance, heating peaks on cold mornings/evenings, while cooling aligns with hot midday hours. This ensured that energy modeling was context-sensitive and highly aligned with regional and seasonal climatic variations. The final output consolidates all factors into realistic yearly profiles that reflect cultural, behavioral, and climatic nuances, setting the stage for further analysis and scenario-based optimization.

\textbf{Prompt Design:} The system prompt highlighted the importance of cultural, seasonal, and weekday/weekend-specific contexts. It required the LLM to assign 24 hourly actions and energy consumption values to each family member, considering interactions among members (e.g., "Father helps Son with homework").

This output highlights the LLM's ability to account for variations in family dynamics, individual schedules, and environmental conditions. During weekends, actions such as 'Family-time' and 'Outdoor-activity' reflect collective family engagements, while weekday routines emphasize individual responsibilities.
\subsection{Validation and Refinement}
To ensure accuracy and realism, outputs were validated against expected cultural norms, weather patterns, and energy usage behaviors. Iterative refinements to the prompts were made based on observed deviations, such as unrealistic activity sequences, energy values, or weather anomalies.

\textbf{Weather Data Refinement:} Initial outputs for weather data revealed two key anomalies:
\begin{enumerate}
    \item \textbf{Excessive Solar Radiation:} In earlier generations, the sum of diffuse and direct solar radiation often exceeded realistic maximum values of 1000 W/m², leading to unrealistically high total solar radiation levels during the midday hours. To address this, the system prompts were refined to enforce a combined maximum limit, ensuring that the sum of diffuse and direct solar radiation remained within plausible meteorological ranges. Additionally, diffuse solar radiation was constrained to always be lower than direct solar radiation during sunny hours.

    \item \textbf{Temperature Timing Misalignment:} Another issue was observed in scenarios such as India during Summer, where the highest temperature of the day occurred at 10 AM instead of the typical late afternoon window (2–4 PM). This anomaly was addressed by explicitly including a directive in the system prompt:

    \begin{quote}
    “\textit{... Peak temperatures typically occur later in the day (2–4 PM), following the peak of solar radiation, due to thermal lag in the ground and air.}”
    \end{quote}
\end{enumerate}
The last modification improved certain responses but did not fully resolve the issue, further motivating the use of externally sourced weather data.

\textbf{Activity Refinement:} In earlier outputs, activities assigned to family members occasionally lacked logical interaction or alignment. For instance, the father was assigned the activity 'Helping the child with homework' while the son was assigned simultaneously 'Playing video games'. Such inconsistencies highlighted the need for additional guidance in the prompts to ensure realistic interactions among family members. To address this, the user prompts were modified to include the directive:
\begin{quote}
“\textit{... Also consider the interactions among family members when assigning actions and consumption values. If the father is with his son helping him with homework, the son’s action must be related to homework, and the father’s action must be related to helping.}”
\end{quote}
This adjustment produced outputs where activities were better synchronized. For instance, in a refined scenario for a family in the USA, the following interactions are observed. The mother is assigned the action 'Helping the son with homework' (0.1 kWh) which reflects her presence in the same room. At the same hour, the son was assigned the action 'Doing homework' (0.1 kWh). The matching consumption values are interpreted as representing shared energy use, divided between both individuals, or as an oversight in the allocation mechanism, highlighting an area for further validation.

\textbf{Refinement Impact:} Following these adjustments, the weather data showed improvements: Most solar radiation values aligned with expected meteorological ranges, enhancing the realism of HVAC energy modeling. Temperature profiles exhibited delayed peaks, reflecting the thermal inertia of the environment. Other prompt refinements included relationships between weather parameters, such as higher temperatures correlating with lower humidity during sunny periods. Surprisingly, when using the \texttt{microsoft/phi-4} LLM model, this interaction became more consistent with real-world observations. Such refinements ensured coherence in the generated activities, energy values, and weather patterns, producing outputs that mirrored realistic household energy dynamics. The iterative process underscores the importance of detailed prompt engineering and continuous validation in leveraging LLMs for synthetic data generation.

\section{Results and Discussion}\label{Sec-Results}
The results generated across all four stages illustrate the efficacy of using LLMs for household energy modeling. This section presents the outcomes, comparing both the \emph{full LLM-based approach} and the \emph{external-weather-data approach}, and discusses their implications for energy consumption modeling, cultural insights, and methodological innovation.

\subsection{Family Structures (Stage 1)}
The family structures synthesized for each country captured both cultural diversity and the unique household dynamics within each region. Regardless of whether the approach used the four LLM-driven stages or external weather data, the same outputs were used to maintain consistency in household definitions when comparing within the same model. 

\textbf{Output Example:}
For instance, using the \texttt{Llama-3.1-405B Instruct} model and for the country of India, the generated family types included a \emph{Joint Family} (Grandfather, Grandmother, Father, Mother, Son, Daughter, Uncle, Aunt, Cousin), illustrating the cultural significance of extended families in Indian society. Appendix~\ref{Sec-Stage1} provides a detailed example of the family types samples. 

All five models successfully generated the requested family structures, albeit with varying degrees of accuracy and consistency. \textbf{\texttt{Llama-3.1-405B-Instruct}} emerged as the most reliable in adhering to cultural and structural prompts, while \textbf{\texttt{Llama-3.3-70B-Instruct}} followed closely, generally adhering to rules but occasionally omitting smaller details. \textbf{\texttt{QwQ-32B-Preview}} produced mostly coherent outputs, though it sometimes failed to comply with prompt requirements. \textbf{\texttt{DeepSeek-R1}} provided valid data but occasionally inserted ages for household members without clear justification. Finally, \textbf{\texttt{phi-4}} struggled to maintain clear parent–grandparent relationships and, instead of distinguishing father, mother, and grandparents individually, sometimes listed them under more generic roles (e.g., “Parents,” “Grandparents”).

\subsection{Weather Range Definition (Stage 2)}
This stage introduces a robust framework for synthetic weather generation, minimizing anomalies such as excessive solar radiation or unrealistic temperature profiles. Because the LLMs are instructed to produce the ranges for the four seasons in a single response, each model ultimately generates six sets of weather ranges (one per country). Appendix~\ref{Sec-Stage2} provides a detailed overview of the seasonal weather parameters in the six selected countries using the \texttt{DeepSeek-R1} model. All five models succeeded at this stage.

\textbf{Output Example:} For the USA in Winter, the temperature range was \mbox{(-10\textdegree C to 10\textdegree C)} and the humidity range was \mbox{(30\% to 80\%)}. These ranges underscore the importance of climatic diversity in shaping energy consumption patterns. Results showed marked seasonal temperature swings for the USA, high summer temperatures in India, low year-round temperatures in Sweden, extreme heat in the UAE, and largely tropical patterns in Brazil.

\subsection{Weather Data Synthesis (Stage 3)}
Hourly weather data are synthesized for each country and season based on the Stage~2 ranges. Although five models were initially tested, only \texttt{Llama-3.1-405B-Instruct}, \texttt{Llama-3.3-70B-Instruct}, and \texttt{phi-4} produced sufficiently consistent and complete hourly data in all queries. By contrast, \texttt{QwQ-32B-Preview} and \texttt{DeepSeek-R1} often returned partial or incomplete outputs, as outlined in \autoref{tab:llms}. 

Despite its lower success rate in fully completing prompts, \texttt{DeepSeek-R1} demonstrated particularly rich reasoning when it \emph{did} generate a valid response, sometimes providing additional contextual clues that could highlight anomalies or inconsistencies in the data. 

To enable a fair comparison with external sources, we also leveraged \texttt{pvlib} to retrieve Typical Meteorological Year data for the capital city of each country, using latitude and longitude coordinates to generate hourly averages. These were then aggregated and formatted to mirror the LLM-based outputs, facilitating smooth integration into our household energy modeling pipeline. For illustrative purposes, four sets of hourly weather profiles for the United Arab Emirates in Summer are provided in Appendix~\ref{Sec-Stage3}. Visual inspection suggests that the external dataset is generally smoother and more physically consistent, making it preferable whenever available. Nonetheless, \texttt{DeepSeek-R1} can serve as a viable alternative for hourly weather synthesis in the absence of reliable external sources. Among those models producing consistent replies, \texttt{phi-4} tended to yield the most realistic seasonal fluctuations when compared against the \texttt{meta-llama} variants. 

\subsection{Energy Consumption Patterns (Stage 4)}
Stage~4 integrates outputs from the earlier stages—culturally grounded family structures (Stage~1) and weather data from either (Stages~2 and~3) or the TMY data—to produce daily household energy consumption profiles. These profiles capture member-specific schedules, HVAC usage, and the interplay between occupant behavior and climate. Weekday consumption often dips during midday when working-age individuals are at work or children are at school, while weekend usage typically increases due to greater home occupancy.

\begin{figure*}[t]
\centering
\subfloat[]{%
    \includegraphics[width=0.90\textwidth]{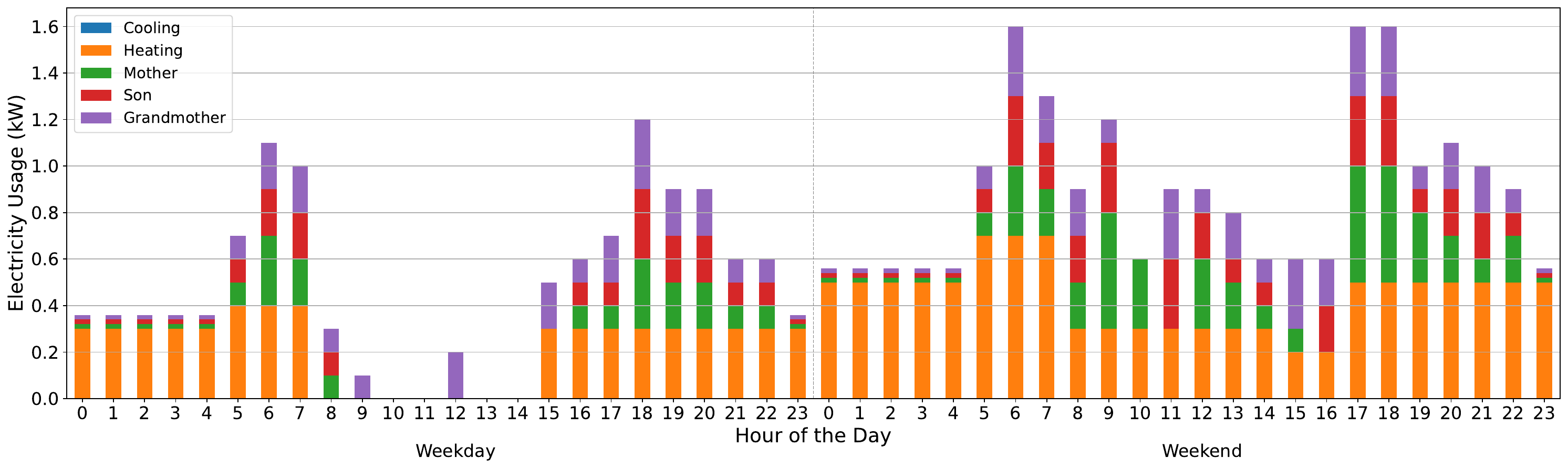}%
    \label{fig:usa_winter}}%
\hfill
\subfloat[]{%
    \includegraphics[width=0.90\textwidth]{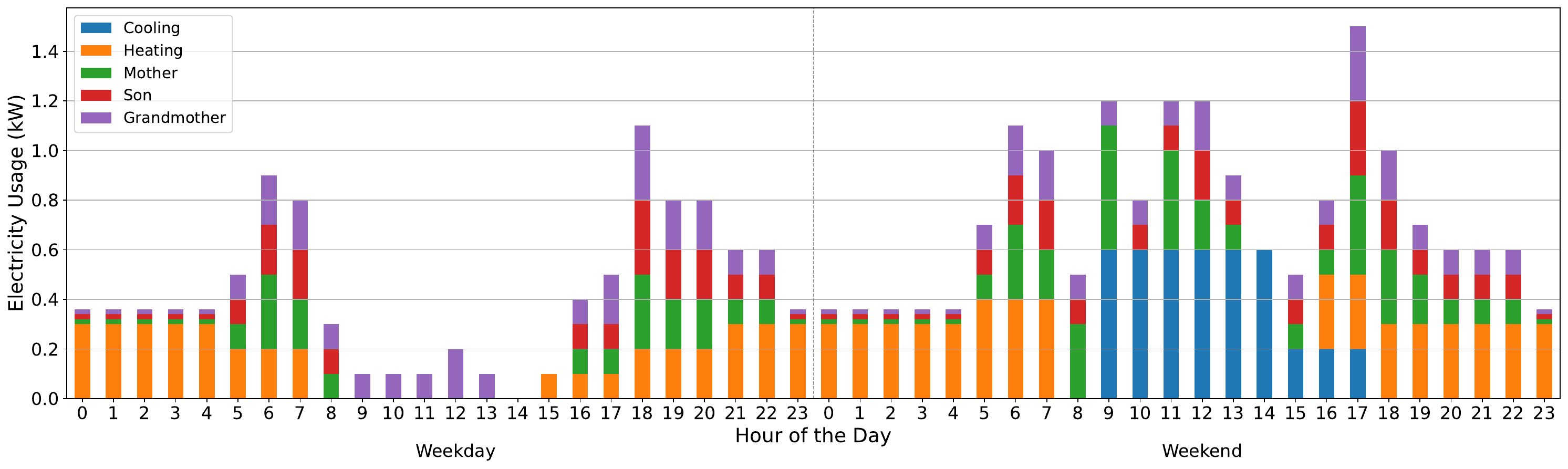}%
    \label{fig:usa_summer}}%
\caption{Hourly energy consumption patterns for a single-parent family in the USA across (a) Autumn and (b) Spring, showing weekday and weekend patterns and using the weather data from Stage 2 and Stage 3, separated by a dashed line to highlight differences in energy consumption for all family members and HVAC actions.}
\label{fig:usa_consumption}
\end{figure*}

\textbf{Output Example:} \autoref{fig:usa_consumption} shows the hourly energy consumption patterns of a single-parent family in the USA during the Autumn and Spring seasons and using the weather data from Stages~2 and~3. The mother and child leave on weekdays, significantly reducing midday consumption until they return in the evening. By contrast, weekends show a higher baseline throughout the day, reflecting continuous occupancy and family-oriented activities. \autoref{fig:japan_consumption} in appendix illustrates another scenario for a nuclear family scenario in Japan using the TMY data.

HVAC usage patterns follow the weather data generated in the previous stages: heating needs peak during colder nighttime and early morning hours, while cooling is activated in warmer seasons to offset midday heat. This synergy between occupant schedules and weather underscores the importance of integrating behavioral and environmental modeling for accurate energy consumption estimates.

\subsection{Discussion and Challenges}
Some challenges persist; for instance, \texttt{Llama-3.3-70B- Instruct} erroneously assigned an energy load of \textbf{0.20\,kWh} for lunch (hour 12) in a joint family scenario in Japan (Autumn), despite the father being at work. This highlights the LLM’s occasional difficulty with context-specific activities, despite comprehensive day-long schedules. Similarly, identical energy values for joint tasks (e.g., ``Helping Son with homework'') suggest a need to refine how collaborative actions are attributed within the consumption model.

However, performance varied among the tested LLMs: \textbf{DeepSeek-R1:} Offered the most realistic weather-stage outputs (Stages~2--3) but struggled with energy consumption modeling (Stage~4). \textbf{Llama-3.1-405B Instruct:} Demonstrated strong consistency and stability across all stages, especially Stages~1 and~4. \textbf{Llama-3.3-70B Instruct:} Generally reliable but prone to small contextual oversights (e.g., misassigning energy loads). \textbf{phi-4:} Served as a solid fallback when DeepSeek-R1 failed to provide complete weather data, though less detailed in its reasoning. \textbf{QwQ-32B-Preview:} Encountered major issues in Stages~3--4, delivering incomplete or inconsistent responses.

A \emph{hybrid approach} could thus leverage:
\begin{itemize}
    \item \textbf{External Sources (e.g., \texttt{pvlib}):} Primary option for weather data when available.
    \item \textbf{DeepSeek-R1:} Secondary choice for generating weather if external data are absent, given its strong performance in Stages~2--3.
    \item \textbf{Meta-Llama-3.1-405B Instruct:} Preferred for family structures Stage~1 and energy patterns Stage~4, offering stable, reliable outputs.
\end{itemize}

In addition to the daily consumption patterns, we assemble country-specific yearly CSV files by incorporating local holidays and weekend definitions, ensuring authentic weekday–weekend distinctions and resulting in continuous, yearlong load profiles that reflect both cultural routines and climate conditions. We use these extended profiles to generate representative energy signatures (see Appendix~\ref{Sec-EngSig}). In the same appendix, we also compare a subset of our generated profiles to the open-source CityLearn Challenge 2022 dataset~\cite{T8/0YLJ6Q_2023}, illustrating how closely our synthetic data align with real-world consumption patterns and demonstrating the potential for robust scenario analysis when integrating synthetic and empirical datasets.

Appendix \ref{Sec-Summary} summarizes the time, number of responses per stage, and total tokens for the LLM. Stages that require granular, hourly outputs (e.g., Stage~4) incur higher computational costs, totaling about 8~hours and 10~minutes for all runs for the \texttt{Meta-Llama-3.1-405B Instruct} model for example. 

\section{Potential Applications}\label{Sec-Potentials}
The results underscore the potential of LLMs to distill complex cultural and contextual knowledge into actionable insights for energy modeling. Key strengths include: \textbf{Scalability:} The framework is readily adaptable to more countries, family types, and weather parameters. \textbf{Realism:} Outputs capture cultural and seasonal nuances, enhancing the credibility of energy simulations. This framework offers versatile applications in household energy modeling and beyond:

\begin{enumerate}
    \item \textbf{Urban Planning \& Infrastructure:}  
    Integrating culturally tailored family structures, synthesized weather data, and energy profiles enables more efficient resource allocation and informs the design of sustainable, energy-efficient cities. The approach can guide renewable energy integration and urban planning under diverse climatic and cultural scenarios.

    \item \textbf{Demand Response \& Grid Optimization:}  
    Generated energy profiles help utilities predict peak loads, optimize demand response, and enhance grid resilience. Detailed HVAC usage insights can shape intelligent home automation and adaptive energy management systems, improving efficiency and reducing costs.

    \item \textbf{Policy Development:}  
    Policymakers can leverage family- and climate-specific usage data to design targeted subsidies, incentives, and regulations grounded in realistic energy needs and consumption patterns.

    \item \textbf{Education \& Research:}  
    The framework provides diverse scenarios for teaching energy modeling, cultural studies, and climate adaptation. It supports investigations into energy equity by simulating usage across socioeconomic contexts, aiding efforts to reduce energy poverty and promote sustainability.
\end{enumerate}

\section{Conclusion}\label{Sec-Conclusion}
The integration of cultural, behavioral, and climatic factors in household energy modeling represents a significant leap forward in understanding and optimizing energy consumption. By introducing a four-stage process that defines weather ranges, synthesizes weather data, and models energy consumption, this research sets the stage for future advancements in sustainable energy management. Continued innovation and collaboration will be key to achieving energy efficiency and equity goals, leveraging this framework as a foundational tool.

Future research should prioritize expanding the geographic and temporal scope by including underrepresented regions and intra-seasonal variations. Incorporating additional cultural parameters, such as socio-economic factors, would provide deeper insights into energy usage patterns and improve cultural sensitivity. These enhancements aim to refine predictions and address biases, ensuring more contextually relevant output.

Advancing the framework further involves enabling real-time simulations to dynamically respond to changes like sudden weather shifts or unexpected occupancy, enhancing its adaptability for disaster response and energy crisis management. Optimizing LLMs specifically for energy modeling through custom datasets could improve contextual accuracy. Scaling the framework to urban and regional models by aggregating household profiles and considering community-level dynamics would also enable broader applications.Ethical considerations remain crucial, particularly in mitigating biases and ensuring compliance with data protection regulations. Transparent and explainable systems are essential as LLMs become increasingly integrated into energy modeling. Interdisciplinary collaboration between energy experts, social scientists, and data scientists can refine methodologies, improve robustness, and ensure meaningful outputs that support equitable and sustainable energy solutions.

\section*{Impact Statement}
This paper presents work whose goal is to advance the field of Machine Learning for energy modeling, potentially accelerating data-driven insights and promoting more sustainable, equitable policy interventions. There are many potential societal consequences of our work, none of which we feel must be specifically highlighted here.

\section*{Acknowledgment}
We acknowledge the usage of icons in \autoref{fig:visual_abstract} designed by various artists from \href{flaticon.com}{flaticon.com}: 6 icons by Freepik, and 1 icon each by monkik, Maxim Basinski Premium, and Awicon.

\bibliography{ref}
\bibliographystyle{icml2025}

%%%%%%%%%%%%%%%%%%%%%%%%%%%%%%%%%%%%%%%%%%%%%%%%%%%%%%%%%%%%%%%%%%%%%
% APPENDIX
%%%%%%%%%%%%%%%%%%%%%%%%%%%%%%%%%%%%%%%%%%%%%%%%%%%%%%%%%%%%%%%%%%%%%
\newpage
\appendix
\onecolumn
\section{Supplementary Experimental Results}
\subsection{Families Types (Sample)}\label{Sec-Stage1}
Each LLM model generated family types per country. Each time the model runs, new family types may be generated. For this reason, when the comparison between using external data or stages 2 and 3 is conducted for the model itself, this stage is run once to ensure consistency. \autoref{tab:family_structures} shows sample family types generated in Stage~1 using the \texttt{Meta-Llama-3.1-405B-Instruct}. 

\begin{table}[h]
\caption{Two examples per country for the family types generated in Stage 1.}
\label{tab:family_structures}
\vskip 0.15in
\begin{center}
\begin{small}
\begin{tabular}{cc}
\toprule
\textbf{COUNTRY} & \textbf{EXAMPLE FAMILY TYPES}\\
\midrule
USA              
& \makecell{'Blended Family',\\'Single-Parent Family'} \\
Japan            
& \makecell{'Traditional Nuclear Family',\\'Three-Generation Family'}\\
India            
& \makecell{'Joint Family',\\'Extended Family with Servant'}\\
Sweden           
& \makecell{'Cohabiting Couple Family',\\'Foster Family'}\\
UAE              
& \makecell{'Polygynous Family',\\'Family with Domestic Worker'}\\
Brazil           
& \makecell{'Multi-generational Family',\\'Family with Adopted Child'}\\
\bottomrule
\end{tabular}
\end{small}
\end{center}
\vskip -0.1in
\end{table}

The outputs reflect each country’s cultural context:
\begin{itemize}
\item \textbf{USA:} A \emph{Blended Family} (Father, Step-Mother, Son, Step-Son, Daughter) illustrates openness to remarriage and step-relations. 
\item \textbf{Japan:} A \emph{Three-Generation Family} (Father, Mother, Son, Daughter, Grandfather, Grandmother) highlights the tradition of including elders in the household. 
\item \textbf{India:} A \emph{Joint Family} (Father, Mother, Son, Daughter, Uncle, Aunt, Cousin) underscores extended familial ties, common in Indian society. 
\item \textbf{Sweden:} A \emph{Cohabiting Couple Family} (Male Partner, Female Partner, Son, Daughter) reflects cultural acceptance of non-marital partnerships. 
\item \textbf{UAE:} A \emph{Family with Domestic Worker} (Father, Mother, Son, Daughter, Domestic Worker) illustrates the region’s reliance on household help. 
\item \textbf{Brazil:} A \emph{Multi-generational Family} (Father, Mother, Son, Daughter, Grandfather, Grandmother, Great-Grandmother) underscores strong intergenerational links, whereas a \emph{Family with Adopted Child} (Father, Mother, Adopted Son, Biological Daughter) reflects acceptance of adoption.
\end{itemize}

\subsection{Weather Ranges (Sample)}\label{Sec-Stage2}
\autoref{tab:weather_ranges} provides seasonal weather parameter ranges across the 6 selected countries which have been generated in Stage 2 using the \texttt{DeepSeek-R1} model.
\subsection{Hourly Weather Parameters (Sample)}\label{Sec-Stage3}
\begin{figure}[t]
\centering
\includegraphics[width=0.78\textwidth]{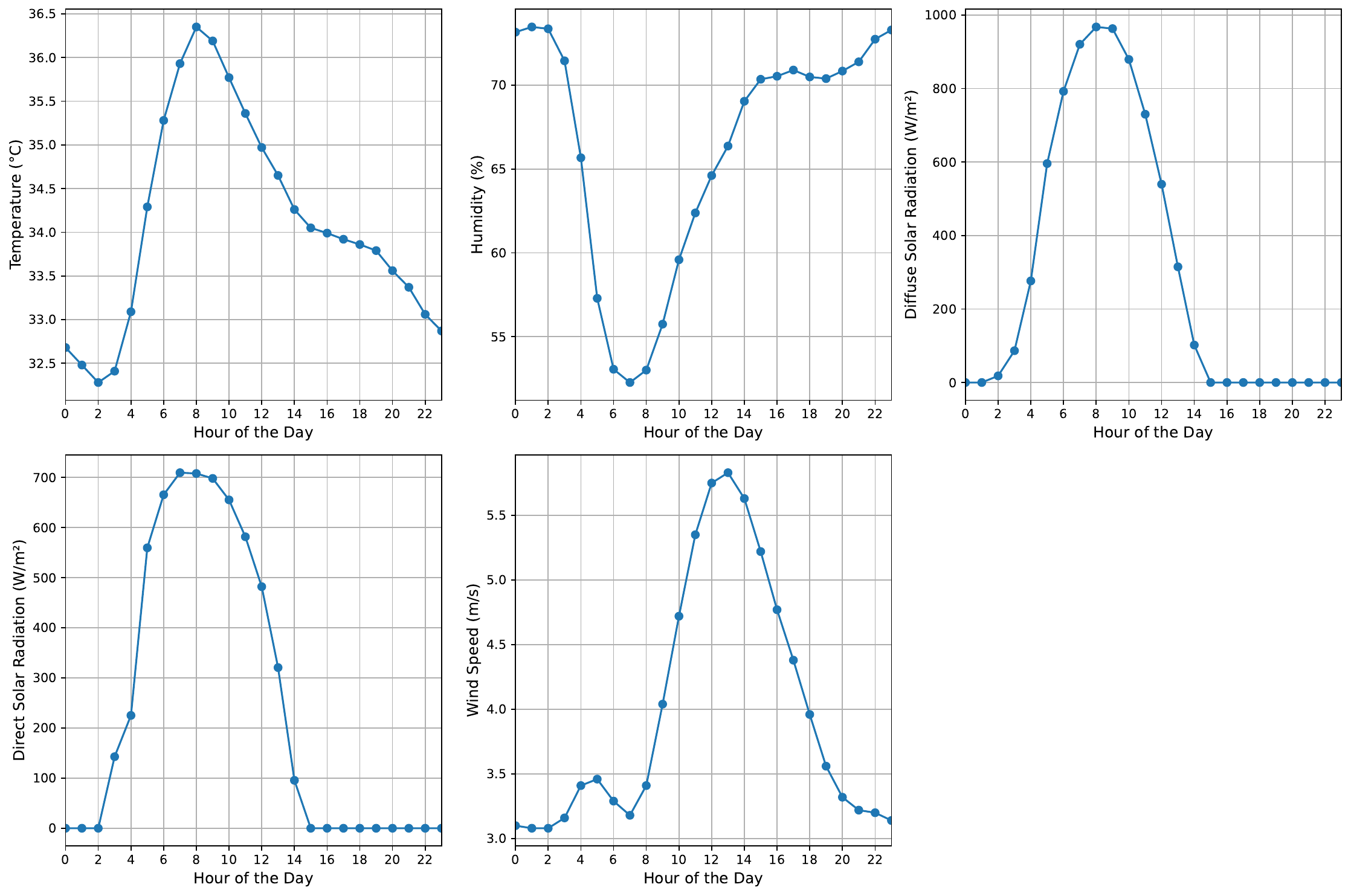}
\caption{Hourly weather values for UAE (Summer) using external TMY data via \texttt{pvlib}.}
\label{fig:uae_summer_tmy}
\vskip -0.2in
\end{figure}

\begin{figure}[ht]
\vskip 0.2in
\centering
\includegraphics[width=0.78\textwidth]{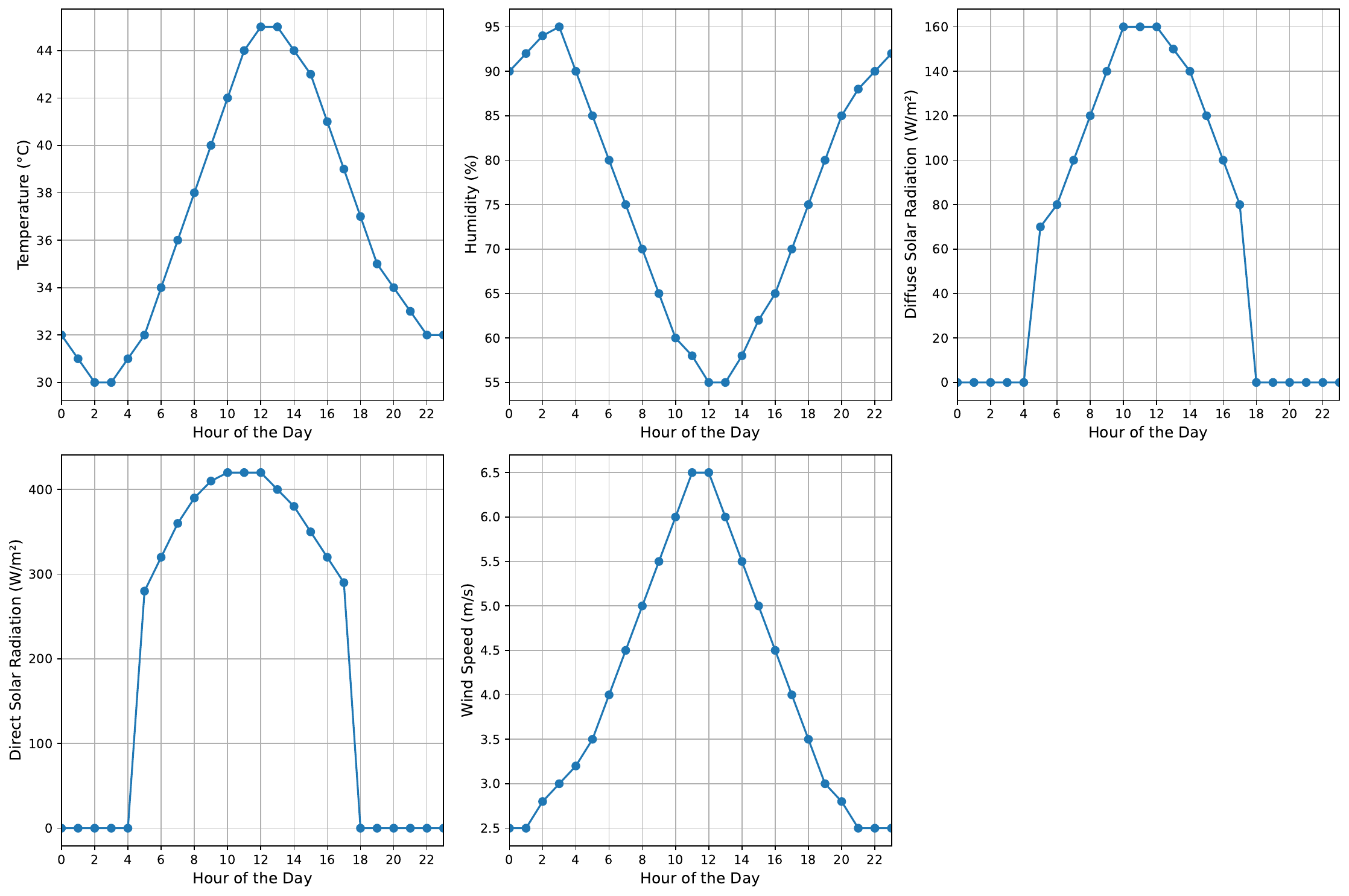}
\caption{Hourly weather values for UAE (Summer) generated by \texttt{DeepSeek-R1}. While occasionally incomplete, the model’s richer reasoning can pinpoint anomalies or inconsistencies.}
\label{fig:uae_summer_deepseek}
\vskip -0.2in
\end{figure}

\begin{figure}[ht]
\vskip 0.2in
\centering
\includegraphics[width=0.78\textwidth]{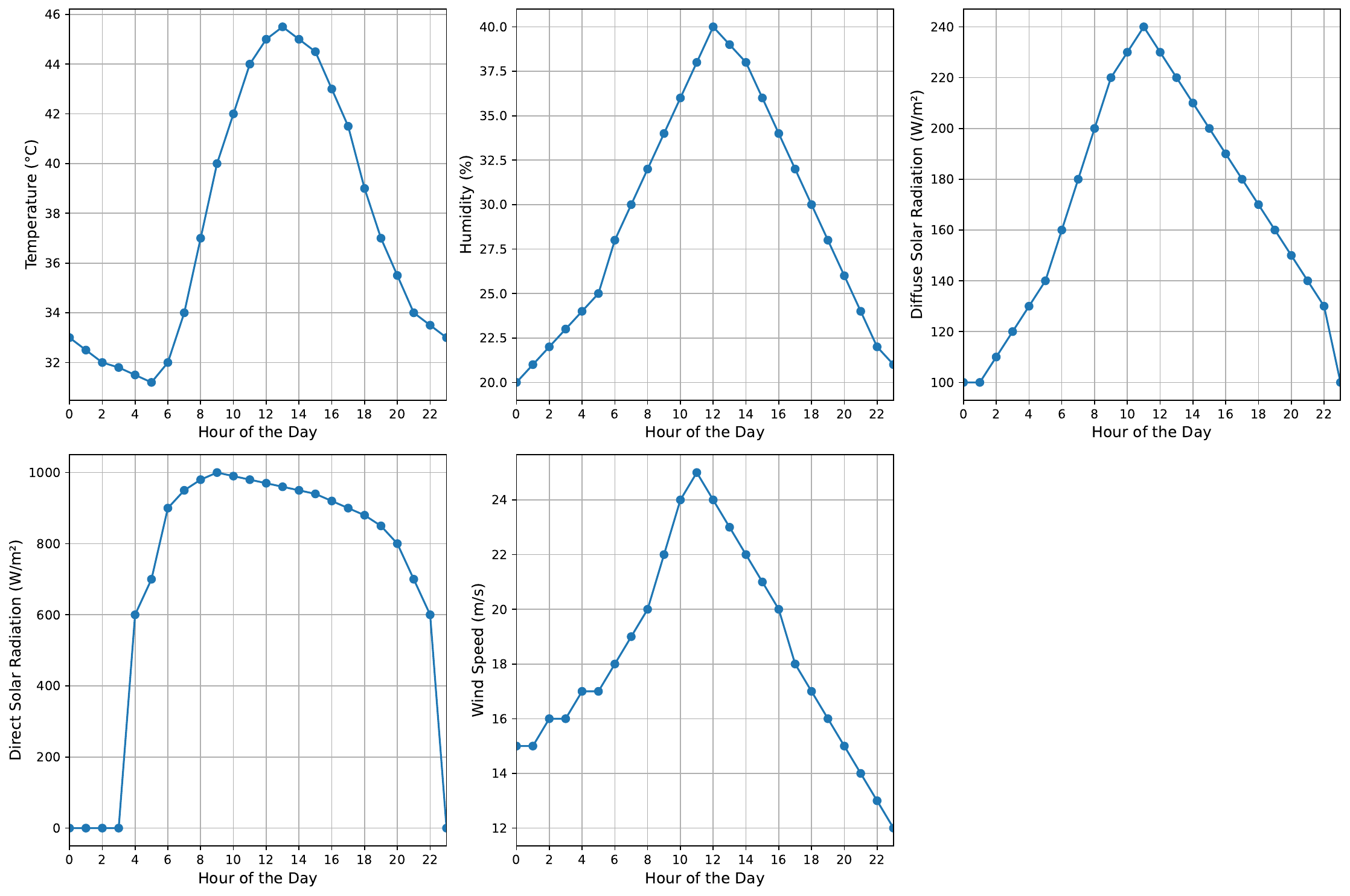}
\caption{Hourly weather values for UAE (Summer) generated by \texttt{phi-4}, which generally provided diurnal patterns. The model did not always minimize humidity during peak temperature intervals, contrary to the prompt’s guidelines.}
\label{fig:uae_summer_phi4}
\vskip -0.2in
\end{figure}

\begin{figure}[ht]
\vskip 0.2in
\centering
\includegraphics[width=0.78\textwidth]{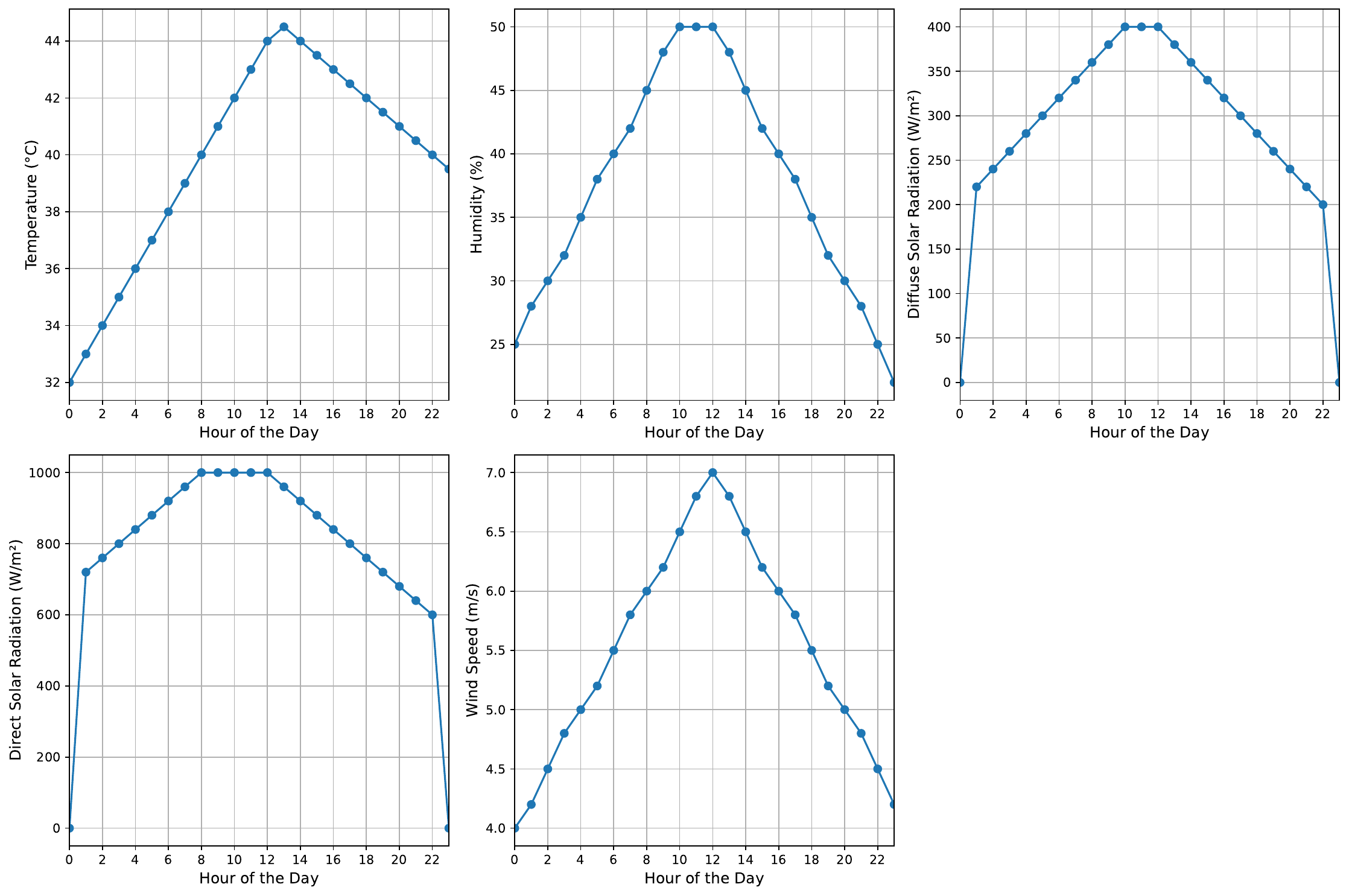}
\caption{Hourly weather values for UAE (Summer) generated by 
\texttt{Meta-Llama-3.1-405B-Instruct}. Although largely coherent, the model did not always minimize humidity during peak temperature intervals, contrary to the prompt’s guidelines.}
\label{fig:uae_summer_llama31}
\vskip -0.2in
\end{figure}

To illustrate the differences between external (TMY) data and various LLM-generated outputs, \autoref{fig:uae_summer_tmy}, \autoref{fig:uae_summer_deepseek}, \autoref{fig:uae_summer_phi4}, and ~\autoref{fig:uae_summer_llama31} present hourly weather parameters for the United Arab Emirates in Summer as generated by: \texttt{pvlib} (TMY external data), \texttt{DeepSeek-R1}, \texttt{microsoft/phi-4}, \texttt{meta-llama/Meta-Llama-3.1-405B-Instruct} respectively. 

\clearpage
\begin{sidewaystable}[t]
\centering
\caption{Seasonal weather parameter ranges across the 6 selected countries which have been generated in Stage 2 using the \texttt{DeepSeek-R1} model.}
\label{tab:weather_ranges}
\resizebox{\textheight}{!}{%
\begin{tabular}{|c|c|c|c|c|c|c|c|c|c|c|c|}
\hline
\textbf{COUNTRY} & \textbf{SEASON} & 
\makecell{\textbf{TEMP.}\\\textbf{MIN}\\($^\circ$C)} & 
\makecell{\textbf{TEMP.}\\\textbf{MAX}\\($^\circ$C)} & 
\makecell{\textbf{HUMIDITY}\\\textbf{MIN}\\(\%)} & 
\makecell{\textbf{HUMIDITY}\\\textbf{MAX}\\(\%)} & 
\makecell{\textbf{DIFFUSE}\\\textbf{SLR. MIN}\\(W/m$^2$)} & 
\makecell{\textbf{DIFFUSE}\\\textbf{SLR. MAX}\\(W/m$^2$)} & 
\makecell{\textbf{DIRECT}\\\textbf{SLR. MIN}\\(W/m$^2$)} & 
\makecell{\textbf{DIRECT}\\\textbf{SLR. MAX}\\(W/m$^2$)} & 
\makecell{\textbf{WIND}\\\textbf{SPD. MIN}\\(m/s)} & 
\makecell{\textbf{WIND}\\\textbf{SPD. MAX}\\(m/s)} \\ \hline
USA & Autumn & 0 & 30 & 40 & 90 & 40 & 150 & 150 & 500 & 2 & 10 \\ \hline
USA & Spring & 5 & 25 & 40 & 85 & 50 & 150 & 200 & 500 & 2 & 10 \\ \hline
USA & Summer & 15 & 40 & 50 & 95 & 80 & 200 & 300 & 800 & 1 & 8 \\ \hline
USA & Winter & -10 & 10 & 30 & 80 & 20 & 100 & 100 & 300 & 3 & 12 \\ \hline
Japan & Autumn & 10 & 25 & 55 & 80 & 70 & 140 & 150 & 350 & 3 & 7 \\ \hline
Japan & Spring & 8 & 20 & 60 & 85 & 80 & 150 & 200 & 400 & 2 & 6 \\ \hline
Japan & Summer & 22 & 35 & 70 & 95 & 100 & 200 & 250 & 600 & 2 & 7 \\ \hline
Japan & Winter & 0 & 12 & 50 & 75 & 50 & 120 & 100 & 250 & 3 & 8 \\ \hline
India & Autumn & 15 & 35 & 60 & 90 & 200 & 500 & 300 & 700 & 1 & 4 \\ \hline
India & Spring & 20 & 40 & 50 & 85 & 150 & 350 & 500 & 900 & 2 & 4 \\ \hline
India & Summer & 25 & 40 & 70 & 95 & 300 & 700 & 200 & 600 & 3 & 6 \\ \hline
India & Winter & 10 & 25 & 40 & 75 & 100 & 300 & 400 & 800 & 1 & 3 \\ \hline
Sweden & Autumn & 5 & 15 & 70 & 90 & 20 & 60 & 80 & 150 & 4 & 9 \\ \hline
Sweden & Spring & 0 & 15 & 65 & 85 & 30 & 80 & 100 & 250 & 3 & 8 \\ \hline
Sweden & Summer & 12 & 25 & 60 & 80 & 50 & 120 & 200 & 350 & 3 & 8 \\ \hline
Sweden & Winter & -10 & 2 & 75 & 95 & 10 & 30 & 20 & 50 & 4 & 10 \\ \hline
United Arab Emirates & Autumn & 25 & 40 & 45 & 90 & 60 & 150 & 200 & 360 & 2.0 & 5.0 \\ \hline
United Arab Emirates & Spring & 22 & 38 & 35 & 75 & 60 & 140 & 220 & 380 & 2.0 & 5.5 \\ \hline
United Arab Emirates & Summer & 28 & 45 & 55 & 95 & 70 & 160 & 280 & 420 & 2.5 & 6.5 \\ \hline
United Arab Emirates & Winter & 15 & 30 & 40 & 85 & 50 & 120 & 180 & 320 & 1.5 & 4.5 \\ \hline
Brazil & Autumn & 20 & 32 & 65 & 90 & 130 & 270 & 280 & 600 & 2 & 5.5 \\ \hline
Brazil & Spring & 20 & 32 & 60 & 88 & 120 & 280 & 300 & 650 & 2 & 5.5 \\ \hline
Brazil & Summer & 22 & 35 & 70 & 95 & 150 & 300 & 200 & 500 & 1 & 5.0 \\ \hline
Brazil & Winter & 18 & 28 & 55 & 85 & 100 & 250 & 350 & 750 & 2 & 6.0 \\ \hline
\end{tabular}
}
\end{sidewaystable}
\clearpage

\clearpage
\subsection{Hourly Energy Consumption Patterns (Sample)}\label{Sec-Stage4}
\begin{figure*}[ht]
\vskip 0.2in
\centering
\subfloat[]{%
    \includegraphics[width=0.95\textwidth]{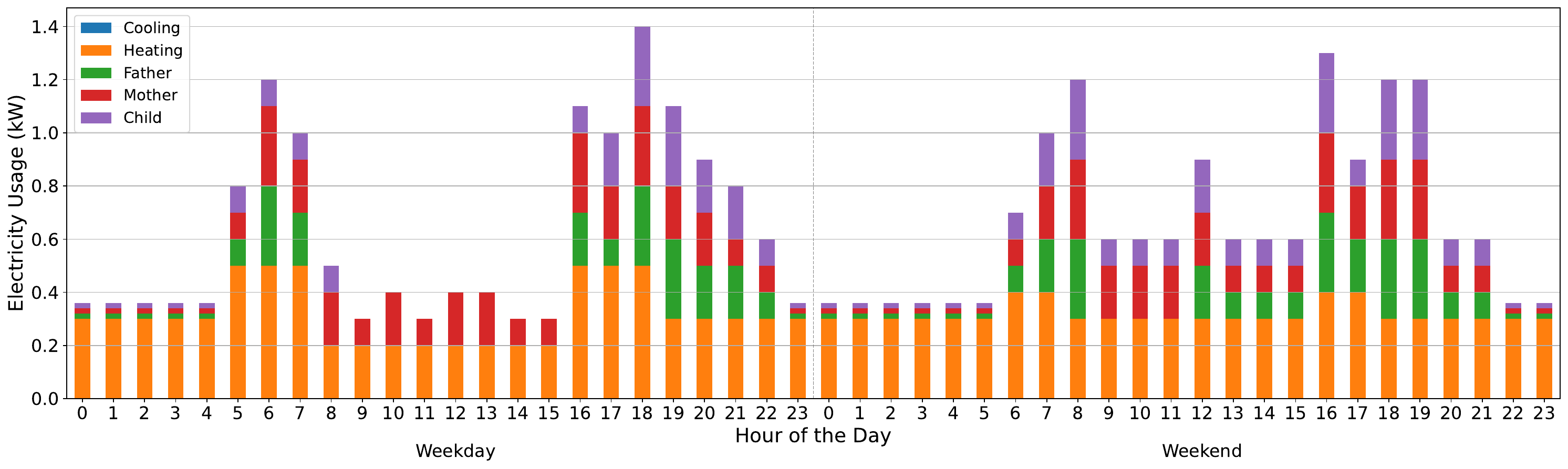}%
    \label{fig:japan_winter}}%
\hfill
\subfloat[]{%
    \includegraphics[width=0.95\textwidth]{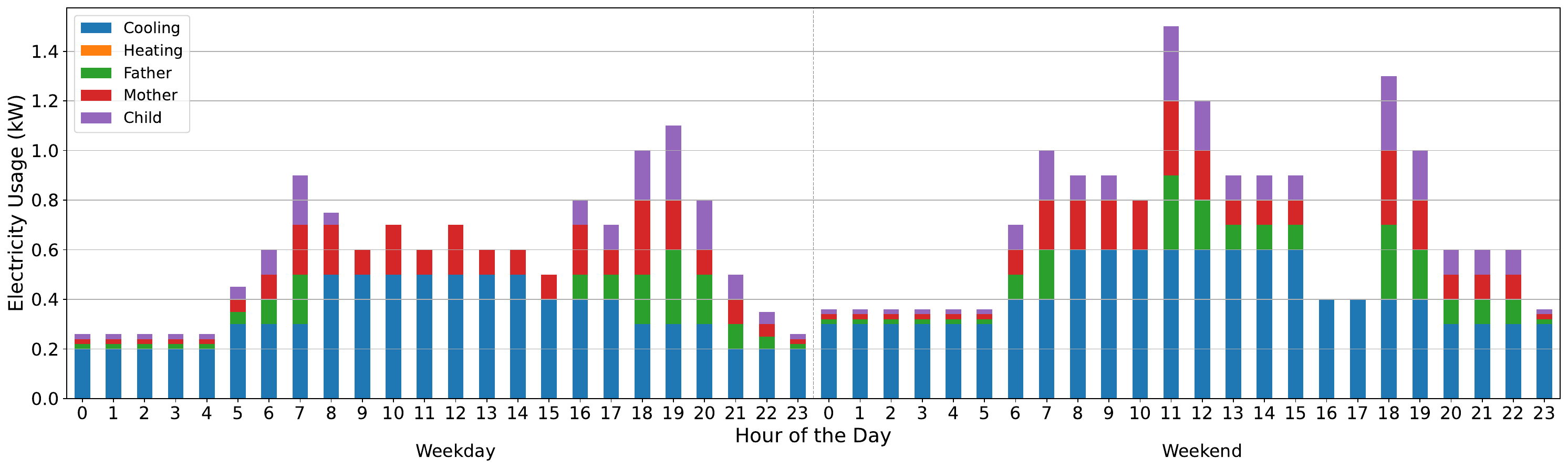}%
    \label{fig:japan_summer}}%
\caption{Hourly energy consumption patterns for a nuclear household in Japan across (a) Winter and (b) Summer and using the TMY data. Dashed lines separate weekdays from weekends, highlighting how both scheduling and climate affect energy usage.}
\label{fig:japan_consumption}
\vskip -0.2in
\end{figure*}

\autoref{fig:japan_consumption} shows the hourly energy consumption patterns for a nuclear household in Japan across Winter and Summer when the TMY data is used for Stage 4. It can be seen in the figure that during the weekday, both the father and the child are not contributing to the electricity usage in the middle of the day. \autoref{tab:energy_usage_family} provides an example of the hourly energy usage of a single-parent family in Sweden during winter for a weekday. These patterns provide insight into how cultural norms, family dynamics, and climatic conditions influence household energy usage in different regions. 

\subsection{Potential Application - Energy Signatures}\label{Sec-EngSig}
Energy signatures are a powerful tool for understanding household energy consumption patterns. By plotting total electricity consumption against external temperature, these signatures provide insights into the household type, behavioral patterns, and thermal characteristics of the building. For example, distinct energy usage profiles can indicate whether a home is well-insulated or poorly insulated, as homes with high heating demand at moderate temperatures likely suffer from heat loss. Additionally, variations in consumption across different temperature ranges help identify occupancy patterns and lifestyle habits. 

\begin{sidewaysfigure}
    \centering
    \includegraphics[width=0.9\textheight]{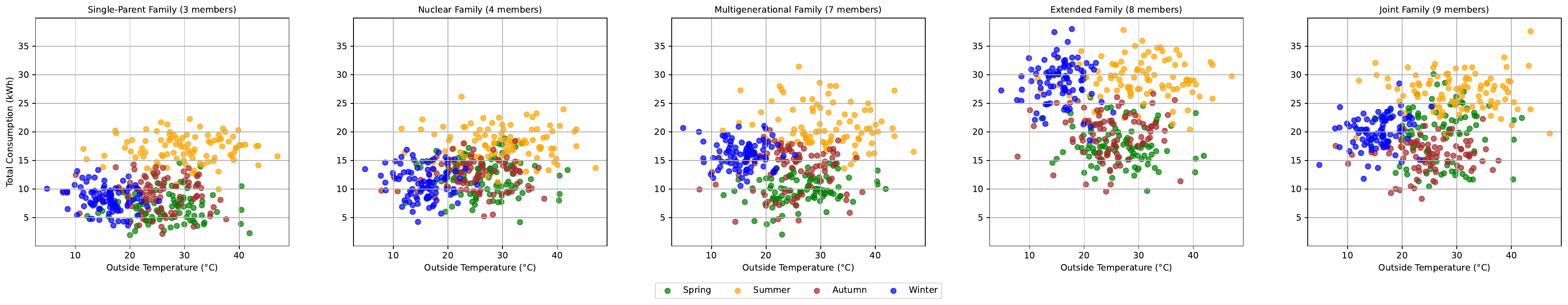}
    \vskip 0.2in
    \includegraphics[width=0.9\textheight]{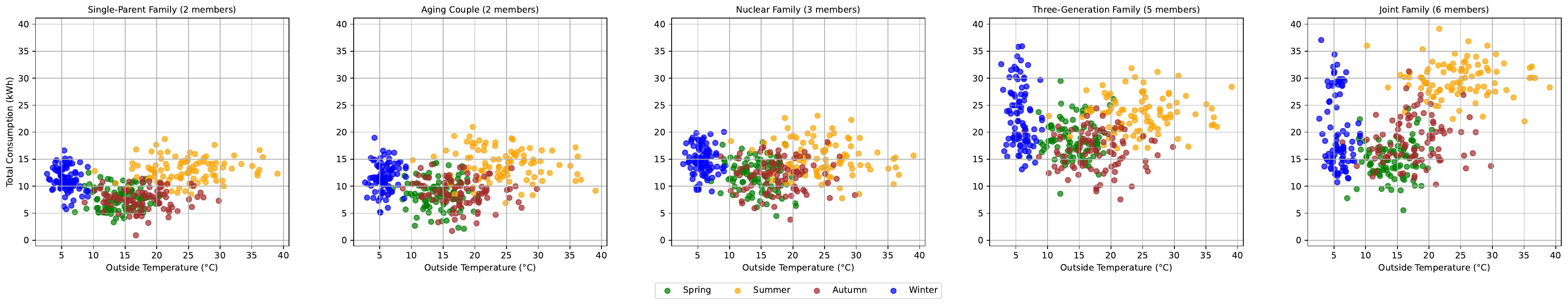}
    \caption{Energy signature comparison for five families in India (top) and Japan (bottom). The x-axis represents outside temperature (retrieved from \texttt{pvlib}), \\while the y-axis shows total electricity consumption per family. Generated by \texttt{Meta-Llama-3.3-70B Instruct}.}
    \label{fig:energy_signature_comparison}
\end{sidewaysfigure}

Energy signature comparisons for five families in India and Japan, generated using TMY data with the \texttt{Meta-Llama-3.3-70B Instruct} model, are shown in \autoref{fig:energy_signature_comparison}. The y-axis shows total electricity consumption per family including the cooling/heating actions. Comparing energy signatures across different regions could allow for cross-cultural analysis of climate adaptation strategies, and heating and cooling efficiency. Such insights can be used to optimize energy efficiency measures, develop demand response strategies, and improve building energy models, making them invaluable for both researchers and policymakers working in smart grid management, energy planning, and sustainability efforts.

\clearpage
\begin{sidewaystable}[t]
\centering
\caption{Example Hourly Energy Consumption Patterns for a Single-Parent Family (Sweden, Winter, Weekday)}
\label{tab:energy_usage_family}
\resizebox{\textheight}{!}{%
\begin{tabular}{|c|c|c|c|c|c|c|c|c|c|}
\hline
\makecell{\textbf{HOUR}} & 
\makecell{\textbf{TOTAL}\\\textbf{ELECTRICITY}\\\textbf{USAGE}} & 
\textbf{MOTHER ACTION} & 
\makecell{\textbf{MOTHER}\\\textbf{CONSUMPTION}} & 
\textbf{SON ACTION} & 
\makecell{\textbf{SON}\\\textbf{CONSUMPTION}} & 
\textbf{HEATING ACTION} & 
\makecell{\textbf{HEATING}\\\textbf{CONSUMPTION}} & 
\textbf{COOLING ACTION} & 
\makecell{\textbf{COOLING}\\\textbf{CONSUMPTION}} \\ 
\hline
0  & 0.34 & Sleeping & 0.02 & Sleeping & 0.02 & \makecell{Nighttime-Heating-\\Maintaining-warmth} & 0.3 & \makecell{No-Cooling-\\Needed-Nighttime} & 0  \\ \hline
1  & 0.34 & Sleeping & 0.02 & Sleeping & 0.02 &   & 0.3 &   & 0  \\ \hline
2  & 0.34 & Sleeping & 0.02 & Sleeping & 0.02 &   & 0.3 &   & 0  \\ \hline
3  & 0.34 & Sleeping & 0.02 & Sleeping & 0.02 & : & 0.3 & : & 0  \\ \hline
4  & 0.34 & Sleeping & 0.02 & Sleeping & 0.02 &   & 0.3 &   & 0  \\ \hline
5  & 0.34 & Sleeping & 0.02 & Sleeping & 0.02 &   & 0.3 &   & 0  \\ \hline
6  & 0.7  & Waking-up & 0.1  & Waking-up & 0.1  & \makecell{Morning-\\Heating-Boost} & 0.5 & \makecell{No-Cooling-\\Needed-Morning} & 0  \\ \hline
7  & 0.9  & Breakfast & 0.2  & Breakfast & 0.2  &   & 0.5 &   & 0  \\ \hline
8  & 0.7  & Getting-ready & 0.3  & Getting-ready & 0.1  & \makecell{Daytime-Heating-\\Maintaining-warmth} & 0.3 & \makecell{No-Cooling-\\Needed-Daytime} & 0  \\ \hline
9  & 0.3  & Commuting & 0    & School & 0    &   & 0.3 &   & 0  \\ \hline
10 & 0.3  & Working & 0    & School & 0    &   & 0.3 &   & 0  \\ \hline
11 & 0.3  & Working & 0    & School & 0    & : & 0.3 & : & 0  \\ \hline
12 & 0.3  & Lunch & 0    & Lunch & 0    &   & 0.3 &   & 0  \\ \hline
13 & 0.3  & Working & 0    & School & 0    &   & 0.3 &   & 0  \\ \hline
14 & 0.3  & Working & 0    & School & 0    &   & 0.3 &   & 0  \\ \hline
15 & 0.5  & Commuting & 0    & Commuting & 0    & \makecell{Afternoon-\\Heating-Boost} & 0.5 & \makecell{No-Cooling-\\Needed-Afternoon} & 0  \\ \hline
16 & 0.7  & Snack & 0.1  & Snack & 0.1  &   & 0.5 &   & 0  \\ \hline
17 & 0.8  & \makecell{Helping-Son-\\with-homework} & 0.2  & Doing-homework & 0.2  & \makecell{Evening-Heating-\\Temperature-drop} & 0.4 & \makecell{No-Cooling-\\Needed-Evening} & 0  \\ \hline
18 & 1    & Dinner & 0.3  & Dinner & 0.3  & : & 0.4 & : & 0 \\ \hline
19 & 0.8  & Relaxing & 0.2  & Relaxing & 0.2  &   & 0.4 &   & 0  \\ \hline
20 & 0.8  & Relaxing & 0.2  & Relaxing & 0.2  &   & 0.4 &   & 0  \\ \hline
21 & 0.7  & Relaxing & 0.2  & Relaxing & 0.2  & \makecell{Nighttime-\\Heating-Stabilizing} & 0.3 & \makecell{No-Cooling-\\Needed-Nighttime} & 0  \\ \hline
22 & 0.5  & Preparing-for-bed & 0.1  & Preparing-for-bed & 0.1  & : & 0.3 & : & 0  \\ \hline
23 & 0.34 & Sleeping & 0.02 & Sleeping & 0.02 &   & 0.3 &   & 0  \\
\hline
\end{tabular}%
}
\end{sidewaystable}
\clearpage

\clearpage

\begin{sidewaysfigure}
    \centering
    \includegraphics[width=0.9\textheight]{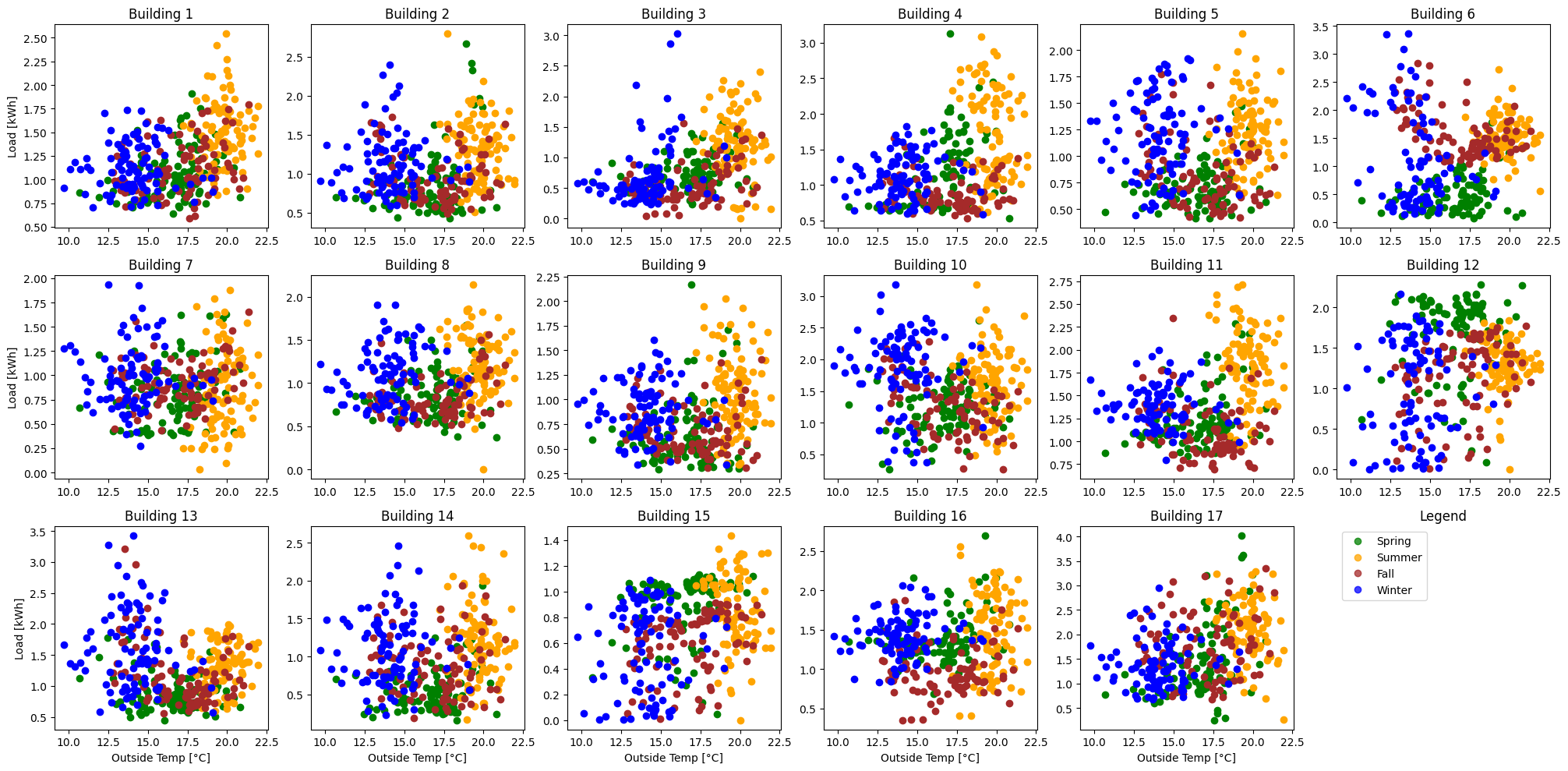}
    \vskip 0.2in
    \caption{Energy signature generated using 17 building profiles in the CityLearn Challenge 2022 dataset~\cite{T8/0YLJ6Q_2023}}
    \label{fig:energy_signature_comparison_citylearn}
\end{sidewaysfigure}
\clearpage

\subsection{Summary of Responses, Usage and Completion Tokens}\label{Sec-Summary}
\autoref{tab:response_summary_405b}, \autoref{tab:response_summary_70b}, \autoref{tab:response_summary_phi}, \autoref{tab:response_summary_qwq}, \autoref{tab:response_summary_dsr1} summarize the duration, number of responses per stage and number of tokens. Note that \autoref{tab:response_summary_qwq} and \autoref{tab:response_summary_dsr1} had some failures in Stage 3 and Stage 4.

\begin{table}[ht]
\caption{Summary of responses, time taken and total tokens per stage for \texttt{Meta-Llama-3.1-405B Instruct}.}
\label{tab:response_summary_405b}
\vskip 0.15in
\begin{center}
\begin{small}
\begin{sc}
\begin{tabular}{cccccc}
\toprule
Stage & \makecell{No. of\\Responses} & \makecell{Avg. Time per\\Response} & \makecell{Total\\Duration} & \makecell{Total Prompt\\Tokens} & \makecell{Total Completion\\Tokens}\\
\midrule
\makecell{Family\\Types}      & 6   & 0:00:32 & 0:03:13 & 2798 & 1821  \\
\makecell{Weather\\Ranges}    & 6   & 0:00:09 & 0:00:59 & 4514 & 1013 \\
\makecell{Weather\\Data}      & 24  & 0:01:09 & 0:27:48 & 47010 & 29302  \\
\makecell{Energy\\Patterns}   & 240 & 0:01:51 & 8:10:36 & 562556 & 508557  \\
\bottomrule
\end{tabular}
\end{sc}
\end{small}
\end{center}
\vskip -0.1in
\end{table}

\begin{table}[ht]
\caption{Summary of responses, time taken and total tokens per stage for \texttt{Meta-Llama-3.3-70B Instruct}.}
\label{tab:response_summary_70b}
\vskip 0.15in
\begin{center}
\begin{small}
\begin{sc}
\begin{tabular}{cccccc}
\toprule
Stage & \makecell{No. of\\Responses} & \makecell{Avg. Time per\\Response} & \makecell{Total\\Duration} & \makecell{Total Prompt\\Tokens} & \makecell{Total Completion\\Tokens}\\
\midrule
\makecell{Family\\Types}      & 6   & 0:00:14 & 0:01:28 & 2798 & 1715  \\
\makecell{Weather\\Ranges}    & 6   & 0:00:08 & 0:00:52 & 4514 & 988 \\
\makecell{Weather\\Data}      & 24  & 0:00:56 & 0:22:42 & 47010 & 31484  \\
\makecell{Energy\\Patterns}   & 240 & 0:01:33 & 6:52:50 & 569745 & 500131  \\
\bottomrule
\end{tabular}
\end{sc}
\end{small}
\end{center}
\vskip -0.1in
\end{table}

\begin{table}[ht]
\caption{Summary of responses, time taken and total tokens per stage for \texttt{phi-4}.}
\label{tab:response_summary_phi}
\vskip 0.15in
\begin{center}
\begin{small}
\begin{sc}
\begin{tabular}{cccccc}
\toprule
Stage & \makecell{No. of\\Responses} & \makecell{Avg. Time per\\Response} & \makecell{Total\\Duration} & \makecell{Total Prompt\\Tokens} & \makecell{Total Completion\\Tokens}\\
\midrule
\makecell{Family\\Types}      & 6   & 0:00:04 & 0:00:24 & 2774 & 1246  \\
\makecell{Weather\\Ranges}    & 6   & 0:00:03 & 0:00:22 & 4490 & 974 \\
\makecell{Weather\\Data}      & 24  & 0:00:20 & 0:08:02 & 44018 & 25563  \\
\makecell{Energy\\Patterns}   & 240 & 0:00:35 & 2:35:05 & 546796 & 482411  \\
\bottomrule
\end{tabular}
\end{sc}
\end{small}
\end{center}
\vskip -0.1in
\end{table}
\clearpage

Note that the last two tables had failures in Stage 3 and Stage 4. The total duration is determined by the number of successful responses shown.
\begin{table}[t]
\caption{Summary of responses, time taken and total tokens per stage for \texttt{QwQ-32B-Preview}.}
\label{tab:response_summary_qwq}
\vskip 0.15in
\begin{center}
\begin{small}
\begin{sc}
\begin{tabular}{cccccc}
\toprule
Stage & \makecell{No. of\\Responses} & \makecell{Avg. Time per\\Response} & \makecell{Total\\Duration} & \makecell{Total Prompt\\Tokens} & \makecell{Total Completion\\Tokens}\\
\midrule
\makecell{Family\\Types}      & 6   & 0:00:05 & 0:00:31 & 2786 & 1361  \\
\makecell{Weather\\Ranges}    & 6   & 0:00:37 & 0:03:42 & 4514 & 10135 \\
\makecell{Weather\\Data}      & 0  & NA & NA & NA & NA  \\
\makecell{Energy\\Patterns}   & 215 & 0:02:34 & 9:14:22 & 534778 & 1329745  \\
\bottomrule
\end{tabular}
\end{sc}
\end{small}
\end{center}
\vskip -0.1in
\end{table}

\begin{table}[t]
\caption{Summary of responses, time taken and total tokens per stage for \texttt{DeepSeek-R1}.}
\label{tab:response_summary_dsr1}
\vskip 0.15in
\begin{center}
\begin{small}
\begin{sc}
\begin{tabular}{cccccc}
\toprule
Stage & \makecell{No. of\\Responses} & \makecell{Avg. Time per\\Response} & \makecell{Total\\Duration} & \makecell{Total Prompt\\Tokens} & \makecell{Total Completion\\Tokens}\\
\midrule
\makecell{Family\\Types}      & 6   & 0:02:32 & 0:15:12 & 2792 & 4587  \\
\makecell{Weather\\Ranges}    & 6   & 0:03:41 & 0:22:11 & 4563 & 6695 \\
\makecell{Weather\\Data}      & 20  & 0:06:59 & 2:19:51 & 36774 & 45223 \\
\makecell{Energy\\Patterns}   & 42  & 0:07:36 & 5:19:39 & 95225 & 105960  \\
\bottomrule
\end{tabular}
\end{sc}
\end{small}
\end{center}
\vskip -0.1in
\end{table}
\clearpage

\section{Prompts Used}\label{prompts}
% \subfile{PromptsUsed}

\subsection{Generating Family Structures}
\begin{mySystem}[]{System Prompt}{}
\begin{lstlisting}
"""You are an expert in energy consumption modeling and forecasting. Your task is to generate realistic family structures across multiple countries, \
considering their cultural and household diversity. 

You can do any reasoning internally, but you MUST NOT share or output your chain-of-thought. You MUST ONLY provide the final JSON answer, \
formatted exactly as specified. Any reasoning or commentary included in your response will be considered an error and invalid. 

**Guidelines**:
1. For the provided country, create unique family types based on cultural norms, ensuring no duplication of family types within the same country.
2. Include family structures like Nuclear Family, Extended Family, Single-Parent Family, Joint Family, etc., relevant to the country's cultural setting.
3. Each family type should have a short clear description and a list of family members that represent this family type.
4. Provide a diverse set of families for the country to represent realistic household scenarios.
5. The language must be English and it should be clear, concise, and culturally sensitive to represent the diversity of family structures across different regions.

**Output Format**:
Once you finish analyzing, respond in valid JSON format, starting with `$$MESSAGE_START$$` and ending with `$$MESSAGE_END$$`. \
Your response must be formatted strictly as follows, without any deviation, commentary, or explanation:

$$MESSAGE_START$$\
[
    {
        "Country": "CountryName",
        "Families": [
            {
                "Family Type": "Type1",
                "Members": ["Member1", "Member2", "Member3"]
            },
            {
                "Family Type": "Type2",
                "Members": ["Member1", "Member2", "Member3", "Member4"]
            },
            ...
        ]
    },
    ...
]\
$$MESSAGE_END$$

**Failure to comply**: If you include reasoning, explanations, or any text outside the JSON format, your response will be invalid and unusable. \
Follow the instructions exactly."""
\end{lstlisting}
\end{mySystem}
\newpage
\begin{myUser}[]{User Prompt}{}
\begin{lstlisting}
"""Generate 5 unique family types for the following country: $COUNTRY$. 

Respond strictly in valid JSON format, starting with `$$MESSAGE_START$$` and ending with `$$MESSAGE_END$$`. \
Do not include any reasoning, explanations, or commentary. Provide only the final JSON result."""
\end{lstlisting}
\end{myUser}
Below is a sample of the using the \texttt{Meta-Llama-3.1-405B-Instruct} response showing 1 country with three family types. Each LLM generated 6 responses which correspond to the 6 countries with 5 family types each. 
\begin{myAssistant}[]{Assistant Prompt}{Sample}
\begin{lstlisting}
[
    {
        "Country": "USA",
        "Families": [
            {
                "Family Type": "Nuclear Family",
                "Members": [
                    "Father",
                    "Mother",
                    "Son",
                    "Daughter"
                ]
            },
            {
                "Family Type": "Single-Parent Family",
                "Members": [
                    "Mother",
                    "Son",
                    "Daughter"
                ]
            },
            {
                "Family Type": "Blended Family",
                "Members": [
                    "Father",
                    "Step-Mother",
                    "Son",
                    "Step-Son",
                    "Daughter"
                ]
            },
        .
        .
        .
]
\end{lstlisting}
\end{myAssistant}

\newpage
\subsection{Generating Weather Data Ranges}
\begin{mySystem}[]{System Prompt}{}
\begin{lstlisting}
"""You are a highly skilled assistant specializing in weather forecasting and energy modeling.
Your task is to provide typical min-max ranges for daily weather parameters across all seasons for a specific country, \
    ensuring the values align with seasonal and geographical contexts.

Follow these guidelines:
1. **Seasonal Weather Data**: Provide min-max ranges for the specified weather parameters for each season in the given country.
2. **Weather Parameters**:
    - Temperature (°C)
    - Humidity (%)
    - Solar Radiation (Diffuse and Direct) (W/m²)
    - Wind Speed (m/s)
3. **Realism**: Ensure that the min-max values align with realistic expectations for each season and the country. Values should reflect:
    - Seasonal variations (e.g., higher temperatures in summer, lower in winter).
    - Regional climatic conditions (e.g., arid regions like the UAE vs. temperate regions like Sweden).
4. **Output Format**: Adhere strictly to the provided structure below to ensure compatibility with downstream processing.

Key Considerations:
1. Use global meteorological data and knowledge to derive the ranges.
2. Avoid extreme outliers unless they represent realistic but rare occurrences.
3. Ensure:
    - The min-max values are realistic for the specified country and season.
    - The ranges reflect typical daily variations within each season.
    - Coherence between the parameters (e.g., high solar radiation corresponds to low humidity).

Once you finish thinking and analysing, the response must start with the '$$MESSAGE_START$$' and ending with '$$MESSAGE_END$$', \
and structured as follows so that I can parse the string and extract the data easily. Your responses must follow the provided \
format without introducing additional text, commentary, or explanations. Any deviation from these guidelines will result \
in invalid parsing. Please provide only the final solution in the specified format without any additional text or explanations.

$$MESSAGE_START$$\
#Temperature#[(Winter,min-value,max-value),(Spring,min-value,max-value),(Summer, min-value,max-value),(Autumn,min-value,max-value)]\
#Humidity#[(Winter,min-value,max-value),(Spring,min-value,max-value),(Summer, min-value,max-value),(Autumn,min-value,max-value)]\
#SolRad-Diffuse#[(Winter,min-value,max-value),(Spring,min-value,max-value),(Summer, min-value,max-value),(Autumn,min-value,max-value)]\
#SolRad-Direct#[(Winter,min-value,max-value),(Spring,min-value,max-value),(Summer, min-value,max-value),(Autumn,min-value,max-value)]\
#Wind-Speed#[(Winter,min-value,max-value),(Spring,min-value,max-value),(Summer, min-value,max-value),(Autumn,min-value,max-value)]\
$$MESSAGE_END$$

**Formatting Rules**:
- Avoid using unescaped single quotes (`'`) within labels. Use a hyphen (`-`) instead (e.g., `Cold-clear`).
- Do not include special characters such as `&`, `*`, `{}`, or newlines (`\\n`, `\\t`) within the labels or other fields. Replace these with a hyphen (`-`) or remove them.
- Ensure there are no extra commas or mismatched quotes in the message.
- Use square brackets `[]` for numeric ranges and ensure proper formatting for seasons.
- Strictly adhere to the format above without additional text, commentary, or explanations."""
\end{lstlisting}
\end{mySystem}
\begin{myUser}[]{User Prompt}{}
\begin{lstlisting}
"""For the country of [$Country$] and in the year of [$Year$], provide typical min-max ranges for the following weather parameters for all seasons:
- Temperature (°C)
- Humidity (%)
- Solar Radiation (split into Diffuse and Direct) (W/m²)
- Wind Speed (m/s)"""
\end{lstlisting}
\end{myUser}
The following is a sample response generated using the \texttt{Meta-Llama-3.1-405B-Instruct} showing the USA weather minimum and maximum ranges per season. Each LLM generated 6 responses, that is one for each country. 
\begin{myAssistant}[]{Assistant Prompt}{Sample}
\begin{lstlisting}
'$$MESSAGE_START$$#Temperature#[(Winter,-20,10),(Spring,-5,25), (Summer,15,35),(Autumn,0,20)]#Humidity#[(Winter,30,70), (Spring,40,80),(Summer,50,90),(Autumn,40,80)]#SolRad-Diffuse#[ (Winter,50,150),(Spring,100,250),(Summer,150,350),(Autumn,100,250) ]#SolRad-Direct#[(Winter,100,300),(Spring,200,500),(Summer,300, 700),(Autumn,200,500)]#Wind-Speed#[(Winter,0,15),(Spring,2,18), (Summer,2,15),(Autumn,2,18)]$$MESSAGE_END$$'
\end{lstlisting}
\end{myAssistant}
\newpage

\subsection{Generating Weather Data}
\begin{mySystem}[]{System Prompt}{}
\begin{lstlisting}
"""You are a highly skilled assistant specializing in weather forecasting and energy modeling. 
Your task is to generate realistic daily weather data for different countries, considering their seasonal, and geographical contexts.
Simulate weather data that aligns with patterns observed in global meteorological datasets for the specified region and season.

Follow these guidelines:
1. **Daily Weather Data**: Generate distinct weather data for a specific country and season, covering 24 hours.
2. **Weather Parameters**:
    - Temperature (°C)
    - Humidity (%)
    - Solar Radiation (Diffuse and Direct) (W/m²)
    - Wind Speed (m/s)
3. **Seasonal Context**: Ensure that the values align with the seasonal conditions and interact realistically \
    (e.g., higher solar radiation typically corresponds to lower humidity during the day, while wind speeds may increase in the afternoon).
4. **Realism**: Values should vary throughout the day to reflect typical weather patterns (e.g., higher solar radiation at noon, lower temperatures at night).
5. **Format**: Adhere strictly to the provided output format, ensuring no extra text or commentary.

Key Considerations:
- The temperature range should reflect the country and season (e.g., colder in winter, hotter in summer).
- Humidity levels should vary realistically based on the time of day and geographic location.
- Solar radiation should differentiate between diffuse and direct components, peaking around noon to early afternoon (12 PM to 3 PM) and falling to zero at night. 
- Ensure that diffuse solar radiation remains below direct solar radiation during sunny hours and that the sum of both aligns with realistic total solar radiation values.
- Peak temperatures typically occur later in the day (2-4 PM), following the peak of solar radiation, due to thermal lag in the ground and air.
- Wind speed should vary naturally throughout the day.

Once you finish thinking and analysing, the response must start with the '$$MESSAGE_START$$' and ending with '$$MESSAGE_END$$', \
and structured as follows so that I can parse the string and extract the data easily. Your responses must follow the provided \
format without introducing additional text, commentary, or explanations. Any deviation from these guidelines will result \
in invalid parsing. Please provide only the final solution in the specified format without any additional text or explanations.

$$MESSAGE_START$$\
#Temperature#[(0, Short-Description, value), (1, Short-Description, value), ..., (23, Short-Description, value)]\
#Humidity#[(0, Short-Description, value), (1, Short-Description, value), ..., (23, Short-Description, value)]\
#SolRad-Diffuse#[(0, Short-Description, value), (1, Short-Description, value), ..., (23, Short-Description, value)]\
#SolRad-Direct#[(0, Short-Description, value), (1, Short-Description, value), ..., (23, Short-Description, value)]\
#Wind-Speed#[(0, Short-Description, value), (1, Short-Description, value), ..., (23, Short-Description, value)]\
$$MESSAGE_END$$

**Formatting Rules**:
- Avoid using unescaped single quotes (`'`) within labels. Use a hyphen (`-`) instead (e.g., `Cold-clear`).
- Do not include special characters such as `&`, `*`, `{}`, or newlines (`\\n`, `\\t`) within the labels or other fields. Replace these with a hyphen (`-`) or remove them.
- Ensure there are no extra commas or mismatched quotes in the message.
- Use square brackets `[]` for lists and ensure all tuples are correctly formatted.
- Strictly adhere to the format above without additional text, commentary, or explanations."""
\end{lstlisting}
\end{mySystem}
\begin{myUser}[]{User Prompt}{}
\begin{lstlisting}
"""For the country of [$Country$] in the year of [$Year$] and during the [$Season$] season, generate a 24-hour weather report covering the following parameters:
Temperature (°C), Humidity (%), Solar Radiation (split into Diffuse and Direct) (W/m²) and Wind Speed (m/s).

Hourly variations reflect typical weather patterns within the following ranges for the country of [$Country$] during the [$Season$] season:
- Temperature: [[$Temperature_Min$], [$Temperature_Max$]] (°C)
- Humidity: [[$Humidity_Min$], [$Humidity_Max$]] (%)
- Solar Radiation (Diffuse): [[$SolRad-Diffuse_Min$], [$SolRad-Diffuse_Max$]] (W/m²)
- Solar Radiation (Direct): [[$SolRad-Direct_Min$], [$SolRad-Direct_Max$]] (W/m²)
- Wind Speed: [[$Wind-Speed_Min$], [$Wind-Speed_Max$]] (m/s)

Ensure that:
1. Values are realistic for the specified country and season, and extreme outliers (e.g., unrealistic highs or lows) are avoided unless representing rare but plausible conditions.
2. Each parameter includes a short descriptive label along with its numeric value e.g., (0,Cold-clear,-2.0).
3. The values for Solar Radiation peak during midday (12 PM to 3 PM), and Temperature peaks later in the day (2-4 PM)."""
\end{lstlisting}
\end{myUser}
Below is a sample response generated using the \texttt{Meta-Llama-3.1-405B-Instruct} model showing the USA weather for a full day per season. Each LLM generated 24 responses at this stage, that is one response per country per season. 
\begin{myAssistant}[]{Assistant Prompt}{Sample}
\begin{lstlisting}
'$$MESSAGE_START$$\n#Temperature#[(0, Cold-clear, -5.0), (1, Cold-clear, -4.5), (2, Cold-clear, -4.0), (3, Very-Cold, -3.5), (4, Very-Cold, -3.0), (5, Chilly, -2.0), (6, Chilly, -1.0), (7, Chilly, 0.0), (8, Brisk, 1.0), (9, Brisk, 2.0), (10, Cool, 3.0), (11, Cool, 4.0), (12, Mild, 5.0), (13, Mild, 6.0), (14, Pleasant, 7.0), (15, Pleasant, 7.5), (16, Warm, 8.0), (17, Warm, 8.0), (18, Cool, 7.0), (19, Cool, 6.0), (20, Chilly, 5.0), (21, Chilly, 4.0), (22, Cold-clear, 3.0), (23, Cold-clear, 2.0)]\n#Humidity#[(0, Low, 40), (1, Low, 42), (2, Low, 45), (3, Moderate, 50), (4, Moderate, 55), (5, Moderate, 60), (6, High, 65), (7, High, 70), (8, High, 68), (9, Moderate, 65), (10, Moderate, 62), (11, Moderate, 60), (12, Low, 58), (13, Low, 55), (14, Low, 50), (15, Low, 45), (16, Low, 40), (17, Low, 38), (18, Moderate, 40), (19, Moderate, 42), (20, Moderate, 45), (21, High, 50), (22, High, 55), (23, High, 60)]\n#SolRad-Diffuse#[(0, No-Sun, 0), (1, Low, 20), (2, Low, 40), (3, Moderate, 60), (4, Moderate, 80), (5, High, 100), (6, High, 120), (7, High, 140), (8, Very-High, 150), (9, Very-High, 145), (10, High, 140), (11, High, 135), (12, Very-High, 130), (13, Very-High, 125), (14, High, 120), (15, High, 115), (16, Moderate, 110), (17, Moderate, 105), (18, Low, 100), (19, Low, 95), (20, Low, 90), (21, No-Sun, 80), (22, No-Sun, 60), (23, No-Sun, 0)]\n#SolRad-Direct#[(0, No-Sun, 0), (1, Low, 50), (2, Low, 100), (3, Moderate, 150), (4, Moderate, 200), (5, High, 250), (6, High, 280), (7, High, 300), (8, Very-High, 290), (9, Very-High, 280), (10, High, 270), (11, High, 260), (12, Very-High, 250), (13, Very-High, 240), (14, High, 230), (15, High, 220), (16, Moderate, 210), (17, Moderate, 200), (18, Low, 190), (19, Low, 180), (20, Low, 170), (21, No-Sun, 160), (22, No-Sun, 120), (23, No-Sun, 0)]\n#Wind-Speed#[(0, Calm, 0.5), (1, Light-Air, 1.0), (2, Light-Breeze, 1.5), (3, Gentle-Breeze, 2.0), (4, Gentle-Breeze, 2.5), (5, Moderate-Breeze, 3.0), (6, Moderate-Breeze, 3.5), (7, Fresh-Breeze, 4.0), (8, Fresh-Breeze, 4.5), (9, Strong-Breeze, 5.0), (10, Strong-Breeze, 5.5), (11, Near-Gale, 6.0), (12, Near-Gale, 6.5), (13, Gale, 7.0), (14, Gale, 7.5), (15, Severe-Gale, 8.0), (16, Severe-Gale, 8.0), (17, Gale, 7.5), (18, Gale, 7.0), (19, Near-Gale, 6.5), (20, Near-Gale, 6.0), (21, Strong-Breeze, 5.5), (22, Strong-Breeze, 5.0), (23, Fresh-Breeze, 4.5)]\n$$MESSAGE_END$$'
\end{lstlisting}
\end{myAssistant}
\newpage
\subsection{Generating Daily Electricity Usage For Families Based on Country and Weather Data}
\begin{mySystem}[]{System Prompt}{}
\begin{lstlisting}
"""You are a highly skilled assistant specializing in energy consumption modeling and forecasting.
Your task is to assist with generating realistic daily electricity usage patterns for families across different countries, considering their cultural, seasonal, and weekday/weekend contexts.
The day type (weekday or weekend) and the season must be considered when selecting the actions and their corresponding consumption values.
For example, the family might have different activities and energy consumption patterns on weekdays compared to weekends but they all must make sense with the cultural, seasonal, and weekday/weekend-specific context.
The sons and daughters might have different activities and energy consumption patterns based on their age and school/work schedules but they cannot be at school on weekends for example.
Also consider the interactions among family members when assigning actions and consumption values. If the father is with his son helping him with homework, the son action must be related to homework and the father action must be related to helping.

Follow these guidelines:
1. **Daily Patterns**: Generate distinct patterns for weekdays and weekends for each family in the selected country and season.
2. **Hourly Actions**: Each family member must have 24 hourly actions, each with a corresponding electricity consumption value.
3. **Interactions**: Include realistic interactions among family members where applicable (e.g., "Father helps Son with homework").
4. **Realism**: Ensure variety and avoid repetition (no action repeated for more than 3 hours, except for sleeping, capped at 8 hours).
5. **Context**: Consider the cultural, seasonal, and weekday/weekend-specific lifestyle patterns when assigning actions and consumption values.
6. **Heating/Cooling**: Include heating or cooling activities based on the season, weather data and members daily usage and include corresponding electricity usage.
7. **Format**: Adhere strictly to the provided output format, ensuring no extra text or commentary.

Key Considerations:
- Electricity consumption values must align with typical appliance usage or activities for each action.
- If a family member is outside the house (e.g., at work, school, or commuting), the corresponding electricity consumption value for that hour must be set to 0.
- Seasonal and cultural factors should influence the actions and electricity usage (e.g., heating during winter, outdoor activities in summer).
- Ensure the output format is clean and structured as requested.
- When the parameters that define the actions are provided, use them as they are and respond accordingly. If they are not provided, you can use your creativity to define them based on the context.

Include:
1. Hourly actions and electricity consumption for each family member.
2. Realistic interactions among family members (e.g., "Father helps Son with homework").
3. Varied actions (no single action repeated for more than 3 hours, except for sleeping, capped at 8 hours).
4. Electricity consumption in kWh for each action, considering appliances/devices used in a typical household.
5. Each family member must have exactly 24 hourly actions and corresponding consumption values.
6. HVAC (Heating, Ventilation, and Air Conditioning) values for heating and cooling, which are influenced by:
   - The total number of members in the house in total and at the current hour.
   - Their activities (e.g., all members sleeping, eating together, or when the house is empty).
   - The weather conditions and time of day.

Ensure variety and align members and HVAC actions with the cultural, seasonal, and weekday/weekend-specific context.

Once you finish thinking and analysing, the response must start with the '$$MESSAGE_START$$' and ending with '$$MESSAGE_END$$', \
and structured as follows so that I can parse the string and extract the data easily. Your responses must follow the provided \
format without introducing additional text, commentary, or explanations. Any deviation from these guidelines will result \
in invalid parsing.

$$MESSAGE_START$$>>>MEMBERS>>>\
#Father#[list of 24 tuples: (hour, action, consumption)]\
#Mother#[list of 24 tuples: (hour, action, consumption)]\
...
#Son#[list of 24 tuples: (hour, action, consumption)]\
>>>HVAC>>>\
#Heating#[list of 24 tuples: (hour, HVAC_action, consumption)]\
#Cooling#[list of 24 tuples: (hour, HVAC_action, consumption)]\
$$MESSAGE_END$$

Example (with dummy data for reference for a family of 3 members (Foster-Father, Foster-Mother, Foster-Daughter) in the USA during Winter on a Weekday, \
with some hours omitted for brevity but ensuring a complete 24-hour representation):

$$MESSAGE_START$$>>>MEMBERS>>>\
#Foster-Father#[(0, Sleeping, 0.02), ...,(12, working, 0), ...,(18, Dinner, 0.3), ...,(23, Sleeping, 0.02)]\
#Foster-Mother#[(0, Sleeping, 0.02), ...,(12, Lunch, 0.2), ...,(18, Dinner, 0.3), ...,(23, Sleeping, 0.02)]\
#Foster-Daughter#[(0, Sleeping, 0.02), ...,(12, Lunch, 0), ...,(18, Dinner, 0.3), ...,(23, Sleeping, 0.02)]\
>>>HVAC>>>\
#Heating#[(0, Nighttime-Heating-Maintaining-warmth, 0.3), ...,(12, No-Heating-Needed-Sunny-day, 0), ...,(18, Evening-Heating-Temperature-drop ,0.3), ...,(23, Nighttime-Heating-Stabilizing, 0.3)]\
#Cooling#[(0, No-Cooling-Needed-Nighttime, 0), ...,(12, Cooling-Afternoon-heat, 0.6), ...,(18, Cooling-Solar-radiation-drop, 0.2), ...,(23, No-Cooling-Needed-Nighttime, 0)]\
$$MESSAGE_END$$

**Important Formatting Rules**:
- Avoid using unescaped single quotes (`'`) within action names. Use a hyphen (`-`) instead (e.g., `Father's help` must become `Father-s help`).
- Do not include special characters such as `&`, `*`, `{}`, or newlines (`\\n`, `\\t`) within the action names or other fields. Replace these with a hyphen (`-`) or remove them.
- Ensure there are no extra commas or mismatched quotes in the message.
- Use square brackets `[]` for lists and ensure all tuples are correctly formatted.
- Strictly adhere to the format above without additional text, commentary, or explanations.

Your responses must strictly follow the format requested above without introducing additional text, commentary, or explanations. \
Any deviation from these guidelines will result in invalid parsing."""
\end{lstlisting}
\end{mySystem}
\begin{myUser}[]{User Prompt}{}
\begin{lstlisting}
"""For a family in [$Country$] in the year of [$Year$], generate their daily electricity usage pattern in the [$Pattern$] considering the season is [$Season$].
The selected family type is [$FamilyType$], which includes the following members: [$Members$] total of [$MembersNum$].

The weather data for the selected country and season are provided for your reference below as csv lists of the 24-hourly values for the \
temperature (°C), humidity (%), direct solar radiation (W/m²), diffuse solar radiation (W/m²), and wind speed (m/s) respectively. \
Use them to determine the heating or cooling actions for the family members.

hour = [$Hour$]
temperature = [$Temperature$] (°C)
humidity = [$Humidity$] (%)
direct_solar_radiation = [$SolarRadiationDirect$] (W/m²)
diffuse_solar_radiation = [$SolarRadiationDiffuse$] (W/m²)
wind_speed = [$WindSpeed$] (m/s)"""
\end{lstlisting}
\end{myUser}
The following is a sample response generated using the \texttt{Meta-Llama-3.1-405B-Instruct} model, illustrating the daily electricity usage pattern for a family in the USA during a winter weekday. Each LLM generated 480 responses, with 20 responses per country per season (10 using the weather data generated from Stage 2 and Stage 3 using the LLM while the other 10 are generated using pvlib weather data replacing the second and third stages. This breakdown includes data for 5 distinct family types for both weekdays and weekends, ensuring comprehensive coverage of various household scenarios across cultural and seasonal contexts.
\begin{myAssistant}[]{Assistant Prompt}{Sample}
\begin{lstlisting}
'$$MESSAGE_START$$>>>MEMBERS>>>#Father#[(0,Sleeping,0.02),(1,Sleeping, 0.02), (2,Sleeping,0.02),(3,Sleeping,0.02),(4,Sleeping,0.02), (5,Waking-up,0.1),(6,Getting-ready,0.2),(7,Breakfast,0.3), (8,Commuting,0),(9,Working,0.1),(10,Working,0.1),(11,Working,0.1), (12,Lunch-break,0.2),(13,Working,0.1), (14,Working,0.1), (15,Commuting,0),(16,Relaxing,0.2),(17,Helping-Son-with-homework, 0.1),(18,Dinner,0.3),(19,Spending-time-with-family,0.2), (20,Watching-TV,0.2),(21,Getting-ready-for-bed,0.1),(22,Sleeping, 0.02),(23,Sleeping,0.02)]#Mother#[(0,Sleeping,0.02),(1,Sleeping, 0.02),(2,Sleeping,0.02),(3,Sleeping,0.02),(4,Sleeping,0.02), (5,Waking-up,0.1),(6,Getting-ready,0.2),(7,Breakfast,0.3), (8,Household-chores,0.2),(9,Household-chores,0.2),(10, Household-chores,0.2),(11,Getting-ready-for-lunch,0.1),(12,Lunch, 0.2),(13,Spending-time-with-Daughter,0.1),(14,Household-chores, 0.2),(15,Getting-ready-for-dinner,0.1),(16,Cooking-dinner,0.3),(17, Spending-time-with-family,0.2),(18,Dinner,0.3),(19,Cleaning-up,0.2) ,(20,Spending-time-with-Father,0.1),(21,Getting-ready-for-bed,0.1), (22,Sleeping,0.02),(23,Sleeping,0.02)]#Son#[(0,Sleeping,0.02), (1,Sleeping,0.02),(2,Sleeping,0.02),(3,Sleeping,0.02),(4,Sleeping, 0.02),(5,Waking-up,0.1),(6,Getting-ready,0.2),(7,Breakfast,0.3), (8,Commuting-to-school,0),(9,At-school,0.1),(10,At-school,0.1), (11,At-school,0.1),(12,Lunch-break,0.2),(13,At-school,0.1),  (14,At-school,0.1),(15,Commuting-from-school,0),(16,Relaxing,0.2), (17,Doing-homework-with-Father-s-help,0.1),(18,Dinner,0.3), (19,Doing-homework,0.1),(20,Watching-TV,0.2), (21,Getting-ready-for-bed,0.1),(22,Sleeping,0.02),(23,Sleeping, 0.02)]#Daughter#[(0,Sleeping,0.02),(1,Sleeping,0.02),(2,Sleeping, 0.02),(3,Sleeping,0.02),(4,Sleeping,0.02),(5,Waking-up,0.1), (6,Getting-ready,0.2), (7,Breakfast,0.3),(8,Commuting-to-school,0),(9,At-school,0.1), (10,At-school,0.1),(11,At-school,0.1),(12,Lunch-break,0.2), (13,At-school,0.1),(14,At-school,0.1),(15,Commuting-from-school,0), (16,Relaxing,0.2),(17,Spending-time-with-Mother,0.1), (18,Dinner,0.3),(19,Doing-homework,0.1),(20,Watching-TV, 0.2),(21,Getting-ready-for-bed,0.1), (22,Sleeping,0.02),(23,Sleeping,0.02)]>>>HVAC>>>#Heating#[
(0,Nighttime-Heating-Maintaining-warmth,0.5), (1,Nighttime-Heating-Maintaining-warmth,0.5), (2,Nighttime-Heating-Maintaining-warmth,0.5), (3,Nighttime-Heating-Maintaining-warmth,0.5), (4,Nighttime-Heating-Maintaining-warmth,0.5), (5,Morning-Heating-Warming-up,0.6), (6,Morning-Heating-Warming-up,0.6), (7,Morning-Heating-Warming-up,0.6), (8,Daytime-Heating-Maintaining-warmth,0.4), (9,Daytime-Heating-Maintaining-warmth,0.4), (10,Daytime-Heating-Maintaining-warmth,0.4), (11,Daytime-Heating-Maintaining-warmth,0.4), (12,Daytime-Heating-Maintaining-warmth,0.4), (13,Daytime-Heating-Maintaining-warmth,0.4), (14,Daytime-Heating-Maintaining-warmth,0.4), (15,Daytime-Heating-Maintaining-warmth,0.4), (16,Afternoon-Heating-Warming-up,0.5), (17,Afternoon-Heating-Warming-up,0.5), (18,Evening-Heating-Maintaining-warmth,0.5), (19,Evening-Heating-Maintaining-warmth,0.5), (20,Evening-Heating-Maintaining-warmth,0.5), (21,Evening-Heating-Maintaining-warmth,0.5), (22,Nighttime-Heating-Maintaining-warmth,0.5), (23,Nighttime-Heating-Maintaining-warmth,0.5)]#Cooling#[(0, No-Cooling-Needed-Nighttime,0),(1,No-Cooling-Needed-Nighttime,0), (2,No-Cooling-Needed-Nighttime,0),(3,No-Cooling-Needed-Nighttime,0) ,(4,No-Cooling-Needed-Nighttime,0),(5,No-Cooling-Needed-Morning,0), (6,No-Cooling-Needed-Morning,0),(7,No-Cooling-Needed-Morning,0), (8,No-Cooling-Needed-Daytime,0),(9,No-Cooling-Needed-Daytime,0), (10,No-Cooling-Needed-Daytime,0),(11,No-Cooling-Needed-Daytime,0), (12,No-Cooling-Needed-Daytime,0),(13,No-Cooling-Needed-Daytime,0), (14,No-Cooling-Needed-Daytime,0), (15,No-Cooling-Needed-Daytime,0), (16,No-Cooling-Needed-Afternoon,0), (17,No-Cooling-Needed-Afternoon,0), (18,No-Cooling-Needed-Evening,0), (19,No-Cooling-Needed-Evening,0), (20,No-Cooling-Needed-Evening,0), (21,No-Cooling-Needed-Evening,0), (22,No-Cooling-Needed-Nighttime,0), (23,No-Cooling-Needed-Nighttime,0)]$$MESSAGE_END$$'
\end{lstlisting}
\end{myAssistant}

\end{document}